\documentclass[11pt]{article}

\usepackage[final]{acl}

\usepackage{times}
\usepackage{latexsym}

\usepackage{booktabs}
\usepackage{ragged2e}
\usepackage{array}
\usepackage{multirow}
\usepackage[table]{xcolor}
\usepackage{graphicx}
\usepackage{amsmath}
\usepackage{amssymb}
\usepackage{mathtools}
\usepackage{amsthm}
\usepackage[most]{tcolorbox}  
\usepackage[dvipsnames]{xcolor}
\usepackage{xcolor}
\usepackage{algorithm}
\usepackage{algorithmic}

\usepackage{hyperref}      
\usepackage{fontawesome5}  
\usepackage{twemojis}      
\usepackage{cuted} 

\definecolor{darkblue}{rgb}{0, 0, 0.5}
\definecolor{unifiedColor}{RGB}{235, 245, 255}

\newcommand{\improve}[1]{\textcolor{orange}{\scriptsize #1}}

\newcommand{\tagUnified}{\hspace{1em}\textit{ + Unified}}

\usepackage[T1]{fontenc}

\usepackage[utf8]{inputenc}

\usepackage{microtype}

\usepackage{inconsolata}

\usepackage{graphicx}

\newcommand{\arrmay}[1]{\textcolor{black}{#1}}
\newcommand{\emnlpcr}[1]{\textcolor{black}{#1}}

\title{Unified-MAS: Universally Generating Domain-Specific Nodes for Empowering Automatic Multi-Agent Systems}

\author{
\bf Hehai Lin\textsuperscript{$^{\spadesuit}$}, ~
Yu Yan\textsuperscript{$^{\spadesuit}$}, ~
Zixuan Wang\textsuperscript{$^{\spadesuit}$}, ~
Bo Xu\textsuperscript{$^{\spadesuit}$}, ~
Sudong Wang\textsuperscript{$^{\spadesuit}$}, ~ \\
\bf
Weiquan Huang\textsuperscript{$^{\spadesuit}$}, ~
Ruochen Zhao\textsuperscript{$^{\diamondsuit}$}, ~
Minzhi Li\textsuperscript{$^{\heartsuit\clubsuit}$}, ~
Chengwei Qin\textsuperscript{$^{\spadesuit}$} \thanks{Corresponding to \href{mailto:chengweiqin@hkust-gz.edu.cn}{chengweiqin@hkust-gz.edu.cn}}~
\\
$^{\spadesuit}$The Hong Kong University of Science and Technology (Guangzhou) \\
$^{\diamondsuit}$Singapore University of Technology and Design ~ $^{\heartsuit}$National University of Singapore \\
$^{\clubsuit}$Institute for Infocomm Research (I$^2$R), A*STAR
}

\begin{document}
\maketitle

\begin{strip} 
\vspace{-5em} 
\begin{center}
\twemoji{brain} \href{https://linhh29.github.io/blog/Unified-MAS/index.html}{Project page} \hspace{0.5em} \textcolor{red}{\faYoutube} \href{https://unified-mas-demo.hehailin.life/}{Gallery \& Live Demo} \hspace{0.5em} \faGithub\ \href{https://github.com/linhh29/Unified-MAS}{Code}
\end{center}
\end{strip} 

\begin{abstract}
Automatic Multi-Agent Systems (MAS) generation has emerged as a promising paradigm for solving complex reasoning tasks. However, existing frameworks are fundamentally bottlenecked when applied to knowledge-intensive domains (e.g., healthcare and law). They either rely on a static library of general nodes like Chain-of-Thought, which lack specialized expertise, or attempt to generate nodes on the fly. In the latter case, the orchestrator is not only bound by its internal knowledge limits but must also simultaneously generate domain-specific logic and optimize high-level topology, leading to a severe architectural coupling that degrades overall system efficacy. To bridge this gap, we propose Unified-MAS that decouples granular node implementation from topological orchestration via offline node synthesis. Unified-MAS operates in two stages: (1) \textbf{Search-Based Node Generation} retrieves external open-world knowledge to synthesize specialized node blueprints, overcoming the internal knowledge limits of LLMs; and (2) \textbf{Reward-Based Node Optimization} utilizes a perplexity-guided reward to iteratively enhance the internal logic of bottleneck nodes. 
Extensive experiments across four specialized domains demonstrate that integrating Unified-MAS into four Automatic-MAS baselines yields a much better performance-cost trade-off, achieving up to a 14.2\% gain while significantly reducing costs. Further analysis reveals its robustness across different designer LLMs and its generalizability to general domains such as mathematical reasoning.
\end{abstract}

\section{Introduction}
\label{sec:introduction}

The rapid evolution of Large Language Models (LLMs) has transformed the landscape of Artificial Intelligence~\cite{ferrag2025llm, yu2026fidesfaithfulinferencedeep, huang2026ama}. Building upon this foundation, LLM-based Multi-Agent Systems (MAS) have emerged as a powerful paradigm, demonstrating superior capabilities by leveraging collaborative intelligence~\cite{lin2025interactive, wu2025furina}. Traditionally, designing effective MAS required meticulous manual engineering by human experts~\cite{wang2022self, shinn2023reflexion}. Recently, the community has experienced a paradigm shift towards automatic MAS generation~\cite{ye2025mas, tran2025multi}. By utilizing techniques such as graph neural networks or code-based optimization, Automatic-MAS can discover novel agentic workflows that often surpass human-designed solutions on general-purpose benchmarks~\cite{ke2025survey}.

\begin{figure*}[htbp]
    \centering
    \includegraphics[width=\linewidth]{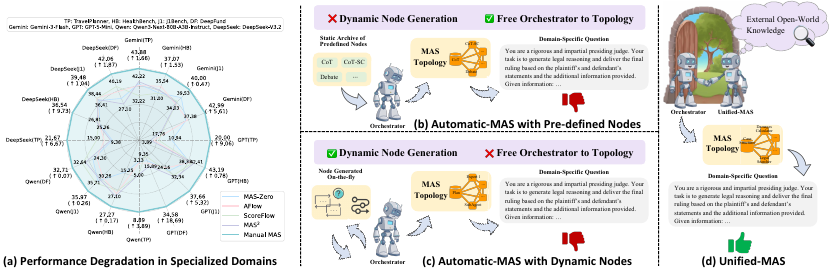} 
    \caption{Overview of MAS paradigms. (a) Performance degradation in specialized domains, where Automatic-MAS with predefined nodes underperforms manual MAS. (b)-(c) Comparison of existing Automatic-MAS paradigms, illustrating the dichotomy between dynamic node generation and topological flexibility. (d) Unified-MAS leverages open-world knowledge to generate domain-specific nodes, effectively empowering existing Automatic-MAS.}
    \label{fig:background}
\end{figure*}

Despite these advancements, a significant limitation persists: the severe performance degradation of Automatic-MAS in \emph{specialized, knowledge-intensive domains}~\cite{hong2023metagpt, xu2025staf}. As illustrated in Figure~\ref{fig:background}(a), our preliminary study reveals that when applied to domains requiring specialized expertise (e.g., legal judgment or clinical diagnosis), they consistently underperform compared to manually crafted, domain-specific MAS. This performance gap stems from the fact that most Automatic-MAS rely on a static set of general-purpose nodes like Chain-of-Thought (CoT) and Debate~\cite{du2024improving}. Lacking specialized priors, the orchestrator tends to merely stack general nodes, failing to capture the nuanced requirements for expert-level tasks~\cite{li2024survey}.

Recent works have attempted to explore dynamic node generation, prompting the orchestrator to invent new sub-agents on the fly~\cite{zhang2025metaagent,ruan2026aorchestra}. However, these approaches suffer from two fundamental flaws. First, they are bound by the \emph{internal knowledge limits} of the LLM. Without grounding in external, domain-specific data (e.g., legal statutes or clinical protocols), the LLM inevitably hallucinates superficial or erroneous node logic~\cite{huang2025survey, yu2026compositioncollapsestablefactual}. Second, it introduces a severe \emph{architectural coupling}. Burdening the orchestrator with the granular implementation of micro-level domain logic distracts and dilutes its primary capability: managing macro-level topology~\cite{ke2026mas}.


To address these challenges, we propose \textbf{Unified-MAS}, a novel framework that advocates for the decoupling of granular node implementation from topological orchestration. As an offline synthesizer, Unified-MAS generates domain-specific nodes for any domain that can be seamlessly integrated into any existing Automatic-MAS. Specifically, Unified-MAS contains two stages: (1) \textbf{Search-Based Node Generation}: Unified-MAS first extracts multi-dimensional keywords from task samples and synthesizes targeted queries. Then, to overcome parametric knowledge limitations, it retrieves external open-world knowledge across diverse sources (i.e., Google, GitHub, and Google Scholar) to distill domain-specific design principles, generating an initial set of specialized nodes.
(2) \textbf{Reward-Based Node Optimization}: Initially generated nodes, while functionally relevant, are often coarse-grained and logically brittle, which may trigger compounding errors in a multi-agent scenario. We introduce a node optimization mechanism driven by a perplexity-guided reward. By quantifying the stability and magnitude of reasoning progress contributed by each node, Unified-MAS identifies \emph{bottleneck nodes} and iteratively refines their internal implementation (e.g., refining prompt constraints or adding necessary sub-agent calls).


We comprehensively evaluate Unified-MAS on four highly specialized benchmarks, i.e., TravelPlanner for constrained travel planning~\cite{xie2024travelplanner}, HealthBench for healthcare~\cite{arora2025healthbench}, J1Bench for legal judgment~\cite{jia2025ready}, and DeepFund for financial decision-making~\cite{li2025time}. We integrate the generated nodes into four general Automatic-MAS baselines, MAS-Zero~\cite{ke2025mas}, AFlow~\cite{zhang2024aflow}, ScoreFlow~\cite{wang2025scoreflow}, and MAS$^2$~\cite{wang2025mas}, and evaluate the system with four different LLMs as orchestrators. The evaluations reveal several key findings:
(1) \emph{Dual Advantage in Performance and Cost.} Unified-MAS consistently drives performance gains, achieving up to a 14.2\% gain, while simultaneously reducing costs. This underscores the critical role of domain-specific priors, positioning our framework as a universal catalyst for elevating general Automatic-MAS into expert-level systems.
(2) \emph{Strong Robustness and Generalizability.} Unified-MAS not only exhibits robust performance across various designer LLMs but also generalizes seamlessly to general domains like mathematics.
(3) \emph{Efficacy of Perplexity-Guided Optimization.} The synthesized nodes progressively improve through reward-based optimization, effectively strengthening their logical reliability in complex domains. Our main contributions are summarized as follows:
\begin{itemize}
    \item We identify the limitations of Automatic-MAS in specialized domains and propose a new paradigm that \emph{decouples} granular node implementation from topology orchestration.
    \item We propose Unified-MAS, which leverages external retrieval to synthesize specialized nodes, and employs perplexity-guided reward optimization to improve their internal logic.
    \item Our extensive experiments demonstrate that Unified-MAS consistently improves the performance of existing Automatic-MAS while reducing costs across complex domains.
\end{itemize}

\section{Related Work}
\label{sec:relatedwork}


\subsection{Automatic-MAS with Pre-defined Nodes}
The most prevalent methods construct Multi-agent Systems (MAS) using a static archive of pre-defined nodes, which consists of manually designed structures, such as CoT, CoT-SC~\cite{wang2022self}, and self-reflection~\cite{madaan2023self, he2025self}, where each node functions as an agent. The orchestrator's role is to determine the optimal topological architecture between these nodes~\cite{chen2024survey, lin2026skill}. 

\textbf{Inference-time} approaches rely on sophisticated prompting and iterative search without updating model weights~\cite{jwalapuram2026illusion}. For example, ADAS represents the MAS as code and iteratively generates new architectures using a Meta Agent Search on a validation set~\cite{hu2024automated}. AFlow employs Monte Carlo Tree Search (MCTS) to discover effective agentic workflows~\cite{zhang2024aflow}, while DyLAN enables multi-round interactions with dynamic agent selection and early-stopping mechanisms to enhance efficiency~\cite{liu2023dynamic}. MAS-Zero introduces a self-reflective feedback loop, allowing the orchestrator to optimize the MAS without requiring an external validation set~\cite{ke2025mas}. \textbf{Training-time} approaches optimize the orchestrator to generate high-quality MAS in one-shot by learning from generated trajectories. ScoreFlow utilizes Score-DPO, a variant of direct preference optimization, to incorporate quantitative feedback into the orchestrator's training~\cite{wang2025scoreflow}. MAS$^2$ learns a self-generative, self-configuring, and self-rectifying workflow~\cite{wang2025mas}, while MAS-Orchestra models MAS construction as a function-calling task optimized via Group Relative Policy Optimization (GRPO)~\cite{ke2026mas}. However, a critical limitation of these methods is their reliance on a static set of general-purpose nodes. As demonstrated in Figure~\ref{fig:background}, when applied to specialized domains, their performance often lags behind manually crafted domain-specific MAS due to the lack of expert knowledge.

\subsection{Automatic-MAS with Dynamic Nodes}
To address the rigidity of pre-defined archives, recent community has turned to dynamic node generation, where the orchestrator attempts to introduce new nodes on the fly based on task requirements. MetaAgent first identifies and implements necessary nodes before optimizing the system using Finite State Machines~\cite{zhang2025metaagent}. EvoAgent serves as a generic method to automatically extend expert agents into MAS via evolutionary algorithms~\cite{yuan2025evoagent}. Similarly, Aorchestra abstracts nodes into a tuple of $\langle \textit{Instruction, Context, Tools, Model} \rangle$, enabling the orchestrator to dynamically populate these slots following task decomposition~\cite{ruan2026aorchestra}. While promising, these approaches are constrained by the orchestrator's internal knowledge. If the necessary domain expertise is absent during the orchestrator's pre-training, the system is prone to hallucinations, resulting in ineffective or erroneous nodes~\cite{ji2023survey}. Furthermore, recent observations suggest that an effective orchestrator should prioritize architectural connectivity rather than the granular implementation of individual nodes~\cite{ke2026mas}.

In this paper, we introduce Unified-MAS, a two-stage workflow designed to generate domain-specific nodes, which can be seamlessly integrated into existing Automatic-MAS frameworks. 
This integration injects essential domain knowledge into the system while liberating the orchestrator from the burden of node design, thereby allowing it to fully leverage its search capabilities to optimize the topological structure of the MAS.

\section{Methodology}
\label{sec:methodology}

\begin{figure*}[htbp]
    \centering
    \includegraphics[width=\linewidth]{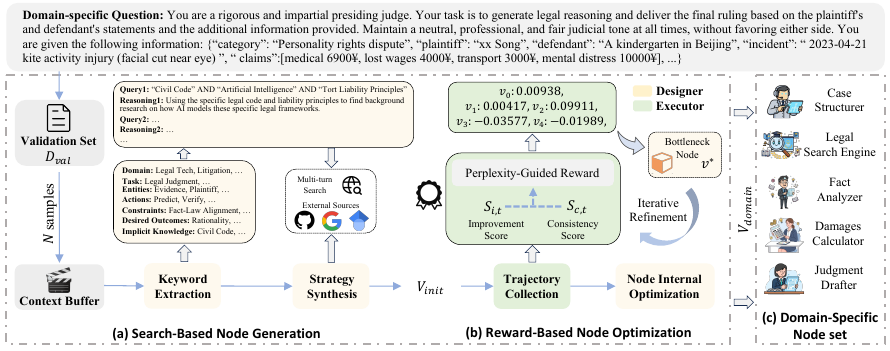} 
    \caption{Illustration of Unified-MAS. (a) Search-Based Node Generation retrieves external knowledge via keyword-strategy driven queries to initialize $V_{init}$. These nodes are subsequently fed into (b) Reward-Based Node Optimization, which iteratively identifies and refines bottleneck nodes guided by a perplexity-based reward. Finally, Unified-MAS generates (c) a domain-specific node set, which can be integrated into existing Automatic-MAS.}
    \label{fig:method}
\end{figure*}

As illustrated in Figure~\ref{fig:method}, Unified-MAS introduces a new paradigm by acting as an offline node synthesizer prior to the Automatic-MAS topological search. This design bridges the gap between general automatic orchestration and domain specificity through a highly decoupled two-stage pipeline: (1) Search-Based Node Generation, which overcomes parametric knowledge limits, and (2) Reward-Based Node Optimization, which improves the internal reasoning logic of individual nodes.


\subsection{Problem Formulation}

Existing Automatic-MAS approaches typically frame the system design as a search problem over a topology space $\Omega$ using a static library of predefined, general-purpose nodes $\mathcal{V}_{fix}$. Let $\mathcal{M}$ represent a MAS configuration defined by its topological structure $\mathcal{G} \in \Omega$ and the selection of functional nodes $V \subseteq \mathcal{V}_{fix}$. The objective is to identify the optimal configuration $\mathcal{M}^*$ that maximizes the performance $\mathcal{R}$ over the data distribution $\mathcal{D}$:
\begin{align}
    \mathcal{M}^* = \mathop{\arg\max}_{\mathcal{G} \in \Omega, V \subseteq \mathcal{V}_{fix}} \mathbb{E}_{x \sim \mathcal{D}} [\mathcal{R}(\mathcal{M}(x; \mathcal{G}, V))]
\end{align}

This formulation inherently limits the solution space to combinations of generic reasoning nodes in $\mathcal{V}_{fix}$. Unified-MAS addresses this limitation by expanding the search space from a static $\mathcal{V}_{fix}$ to a dynamically domain-adaptive set $\mathcal{V}_{domain}$.

\subsection{Search-Based Node Generation}

\paragraph{Multi-Dimensional Keyword Extraction.}
To construct $\mathcal{V}_{domain}$, we first sample $N$ examples from a validation set $\mathcal{D}_{val}$ to form a context buffer $\mathcal{C}$. We prompt the LLM to analyze $\mathcal{C}$ and extract keywords across seven dimensions. This granular decomposition ensures no critical aspect of the domain is overlooked. (1) \textit{Domain}: the macro-industry context (e.g., Fintech); (2) \textit{Task}: the core technical problem (e.g., decision-making); (3) \textit{Entities}: the specific data entities such as company news; (4) \textit{Actions}: the operations or methods performed on these entities; (5) \textit{Constraints}: task requirement such as low latency; (6) \textit{Desired Outcomes}: the target metrics (e.g., accuracy); and (7) \textit{Implicit Knowledge}: latent expert intuitions that are not explicitly stated but are essential for success.

\paragraph{Strategy-Driven Query Synthesis.}
We then synthesize these seven dimensions into four targeted search strategies, each designed to retrieve a specific layer of system design knowledge:
(1) Strategy A (\textbf{Background Knowledge}): combining \textit{Domain} and \textit{Implicit Knowledge} to retrieve background information and survey papers;
(2) Strategy B (\textbf{System Architecture}): combining \textit{Task} and \textit{Constraints} to search for architectural patterns that satisfy specific requirements;
(3) Strategy C (\textbf{Code Implementation}): combining \textit{Entities} and \textit{Actions} to locate repositories for libraries handling specific data types from GitHub;
and (4) Strategy D (\textbf{Evaluation}): combining \textit{Task} and \textit{Desired Outcomes} to identify standard benchmarks and evaluation metrics for this specific domain.

\paragraph{Knowledge Aggregation and Node Generation.}
Finally, we perform multi-turn search~\cite{zhao2026training} using appropriate search engines, and aggregate the retrieved content into strategy-specific summaries. Based on these summaries and guided by a node generation prompt, the LLM generates an initial node set $\mathcal{V}_{init} = \{v_1, \dots, v_m\}$, where each node $ v_i$ represents a domain-specific agent including its system prompts and tool specifications.

\subsection{Reward-Based Node Optimization}

Although the initial nodes in $\mathcal{V}_{init}$ successfully capture essential domain priors, possessing knowledge does not equal robust reasoning. The preliminary nature of their generation often leaves their internal implementation superficial, struggling to handle the nuanced logic required for expert-level tasks. Without iterative refinement, these unstable reasoning mechanics can easily bottleneck the overall system efficacy. Therefore, to transition these nodes from coarse blueprints into reliable operators, we formulate MAS execution as a trajectory reasoning, assign a reward for each node, and optimize the \emph{bottleneck node} with the lowest reward.

Although some nodes are logically parallel, their outputs can be treated as being sequentially appended to the MAS output during execution. Let a reasoning trajectory be a sequence of states $\tau = \{h_0, h_1, \dots, h_m\}$ generated by the sequential execution of nodes $\{v_1, \dots, v_m\}$. Here, $h_0$ represents the empty context before any node execution, while $h_t$ (for $t \ge 1$) denotes the output generated by node $v_t$. The accumulated context after executing node $v_t$ is defined as the concatenation of all preceding outputs: $A_t = [h_0, h_1, \dots, h_t]$.

\emnlpcr{
Dynamically generated subtasks rarely have gold-standard intermediate labels, making it difficult to obtain reliable dense rewards for individual sub-agents. Terminal correctness offers only sparse, execution-level supervision and cannot indicate whether a given node advances or detracts from the reference solution. We therefore use perplexity as a proxy reward.
}
Specifically, to evaluate the effectiveness of each node, we measure how well the accumulated reasoning trajectory predicts the ground-truth answer $y$. We compute the perplexity of generating $y$ given the input question $q$ and the accumulated context $A_t$ under an LLM $P_{\theta}$:
\begin{align}
    \text{PPL}(y | q, A_t) = \exp( -\frac{1}{|y|} \sum_{j=1}^{|y|} \log P_{\theta}(y_j | q, A_t) )
\end{align}

Based on this definition, we derive an objective function $\mathcal{J}$ by maximizing the negative log-perplexity, which reflects the predictability of the answer $y$ given the accumulated reasoning steps:
\begin{align}
\mathcal{J}(P_\theta, y, q, A_t) &= -\log(\text{PPL}(y | q, A_t))  \nonumber\\  &= \frac{1}{|y|} \sum_{j=1}^{|y|} \log P_{\theta}(y_j | q, A_t)
\end{align}
A higher $\mathcal{J}$ corresponds to lower perplexity, indicating that the sequence of reasoning steps up to node $v_t$ has effectively reduced the model's uncertainty and guided the system closer to the correct solution. To standardize evaluation across different queries, we define $\mathcal{J}_0$ as the predictability of the answer using the model's direct inference capability, i.e., with an empty context $A_0$, $\mathcal{J}_0=\mathcal{J}(P_\theta, y, q)$.




To optimize nodes based on the objective defined above, we evaluate each node from two complementary perspectives: \emph{utility} and \emph{stability}. An effective node should provide a reasoning path that is not only impactful (yielding a considerable gain) but also consistent (avoiding erratic fluctuations)~\citep{liu2025rectifying}. We therefore introduce two scores to assess the quality of node $v_t$:


\paragraph{Improvement Score ($\mathcal{S}_{i, t}$)} It measures the relative gain in the objective compared to the baseline $\mathcal{J}_0$, reflecting the strength of the node's contribution. Formally,
\arrmay{
\begin{align}
\mathcal{S}_{i, t} = \tanh(\delta(P_\theta, y, q, A_t) + 1) \\
\delta(P_\theta, y, q, A_t) =  \frac{\mathcal{J}(P_\theta, y, q, A_t) - \mathcal{J}_0}{\left |\mathcal{J}_0\right | } 
\end{align}
}
where $\delta(P_\theta, y, q, A_t)$ represents the normalized improvement over direct inference. The $\tanh$ function is used to smooth outliers and bound the score.

\paragraph{Consistency Score ($\mathcal{S}_{c, t}$)} It assesses the stability of the reasoning process. To measure whether the benefit improves consistently as reasoning depth increases, we compute Kendall's Tau correlation coefficient~\cite{kendall1938new} between the sequence of objective values $\{\mathcal{J}_1, \dots, \mathcal{J}_t\}$ and their corresponding step indices. The consistency score is:
\begin{align}
\mathcal{S}_{c, t} = \frac{2}{t(t-1)} \sum^{i<j}_{1 \le i,j \le t} \text{sgn}(\mathcal{J}_i - \mathcal{J}_j) \cdot \text{sgn}(i-j)
\end{align}
where $\text{sgn}(\cdot)$ denotes the Signum function. A higher $\mathcal{S}_c$ indicates a more stable reasoning.

The \textbf{Node Quality Score ($\mathcal{S}_t$)} is computed as a weighted combination of these two scores:
\begin{align}
\mathcal{S}_t = (1-\alpha) \mathcal{S}_{i, t} + \alpha \mathcal{S}_{c, t}
\end{align}
where $\alpha$ is a balancing hyperparameter. Based on this score, we define the perplexity-guided reward of node $v_t$ as \emph{the incremental gain in node quality}:
\begin{align}
r_t = \begin{cases}
    \mathcal{S}_t - \mathcal{S}_{t-1} & \text{if } t > 1, \\
    \mathcal{S}_t                     & \text{if } t = 1
\end{cases}
\end{align}

To refine node implementations, we perform optimization for $K$ epochs on the validation set $\mathcal{D}_{val}$. In each epoch, we calculate the average reward $\bar{r}(v)$ for each node $v \in \mathcal{V}_{init}$ across all samples of $\mathcal{D}_{val}$. The node with the lowest average reward is identified as the \emph{bottleneck node}:
\begin{align}
v^* = \mathop{\arg\min}_{v \in \mathcal{V}_{init}} \bar{r}(v)
\end{align}

We then retrieve the samples where $v^*$ produces the lowest rewards and use them to refine its internal instructions or add additional LLM calls to maximize future rewards. Importantly, in each epoch, samples for which $v^*$ is not the lowest-reward node are excluded from the optimization process, ensuring targeted and stable refinement.

There are two types of LLMs in Unified-MAS. To distinguish them from the LLMs used in Automatic-MAS (orchestrator), we denote them as Designer and Executor. The Designer is responsible for generating and optimizing domain-specific nodes. We employ Gemini-3-Pro as the default Designer due to its strong capabilities. The effect of different Designer models is further investigated in Section~\ref{sec:different designers}. The Executor executes nodes and collects trajectories to compute the perplexity-guided reward. Considering that this computation requires direct access to token-level logits and the practical deployment, we employ Qwen3-Next-80B-A3B-Instruct as the default Executor.



\section{Experimental Settings}
\label{sec:experiments}

\paragraph{Benchmarks and Evaluation Metrics.}
We select four benchmarks spanning different specialized domains.
(1) \textbf{TravelPlanner}~\cite{xie2024travelplanner} for constrained planning. Performance is measured by accuracy. 
(2) \textbf{HealthBench}~\cite{arora2025healthbench} for health diagnosis. Responses are scored against a rubric using an LLM-Judge.
(3) \textbf{J1Bench}~\cite{jia2025ready} simulates automatic legal adjudication. The agent synthesizes conflicting testimonies to produce a final verdict, evaluated by an LLM-Judge under a unified standard. 
(4) \textbf{DeepFund}~\cite{li2025time} for stock market decision-making and evaluated by accuracy. 
All metrics are normalized to $[0, 100\%]$. We report the average performance and the average cost (in USD \$).
The dataset statistics are provided in Appendix~\ref{appendix:statistics of benchmarks}.
\arrmay{We elaborate the retrieval leakage control in Appendix~\ref{appendix:leakage}.}
The LLM-as-a-Judge prompts are cataloged in Appendix~\ref{appendix:prompt details}.


\paragraph{Baselines.}
We adopt three categories of MAS to ensure a comprehensive evaluation.
(1) \textbf{Specific Manual MAS}: PMC~\cite{zhang2025planning} for TravelPlanner, Diagnosis-MAS~\cite{chen2025enhancing} for HealthBench, Court-MAS~\cite{jia2025ready} for J1Bench, and DeepFund-MAS~\cite{li2025time} for DeepFund. These serve as the manual-design performance standard.
(2)  \textbf{Automatic-MAS with Dynamic Nodes}: 
MetaAgent~\cite{zhang2025metaagent}, EvoAgent~\cite{yuan2025evoagent}, and AOrchestra~\cite{ruan2026aorchestra}, which generate nodes on the fly during problem solving.
(3)  \textbf{Automatic-MAS with Pre-defined Nodes}: We benchmark against leading Automatic-MAS that rely on static nodes, i.e., AFlow~\cite{zhang2024aflow}, MAS-Zero~\cite{ke2025mas},  ScoreFlow~\cite{wang2025scoreflow}, and MAS$^2$~\cite{wang2025mas}. We empower these baselines by replacing their general nodes with the Unified-MAS's domain-specific node libraries.

\begin{table*}[ht]
    \centering
    
    \resizebox{\textwidth}{!}{%

    \begin{tabular}{l | ccccc c | ccccc c}
    \toprule
    \multirow{2.5}{*}{\textbf{Method}} & \multicolumn{6}{c|}{\textbf{Gemini-3-Flash}} & \multicolumn{6}{c}{\textbf{GPT-5-Mini}} \\
    \cmidrule(lr){2-7} \cmidrule(lr){8-13}
     & TP & HB & J1 & DF & Avg.Perf $\uparrow$ & Avg.Cost $\downarrow$ & TP & HB & J1 & DF & Avg.Perf $\uparrow$ & Avg.Cost $\downarrow$ \\
    \midrule
    
    Vanilla      & 38.33 & 26.36 & 33.25 & 16.82 & 28.69 & 2.484 & 3.89 & 37.84 & 23.77 & 12.15 & 19.41 & 0.469 \\
    
    Manual MAS        &43.88 & 37.07 & 40.00 & 42.99 & 40.99 & 21.898& 20.00 & 43.19 & 37.66 & 34.58 & 33.86 & 17.164\\
    MetaAgent    &41.11 & 29.67 & 37.40 & 19.63 & 31.95 & 4.116& 2.22 & 39.04 & 30.39 & 13.08 & 21.18 & 1.443\\
    EvoAgent   & 41.11 & 34.13 & 41.82 & 37.38 & 38.61 & 44.791& 3.89 & 41.40 & 36.49 & 12.15 & 23.48 & 17.498\\
    AOrchestra   &40.00 & 28.98 & 34.16 & 22.43 & 31.39 & 6.856& 6.67 & 38.41 & 30.00 & 17.76 & 23.21 & 3.131\\
    \midrule
    
    MAS-Zero   &40.61 & 31.30 & 39.53 & 35.94 & 36.85 & 132.179 & 10.94 & 38.33 & 26.72 & 12.50 & 22.12 & 111.910\\
    \rowcolor{unifiedColor}
    \tagUnified  & 46.88 & 33.60 &\textbf{48.91} & 48.44 & 44.46 \improve{+7.61} & 123.803 \improve{-8.376}& \textbf{23.44} & 40.21 & 30.94 & 28.12 & 30.68 \improve{+8.56} & 44.011 \improve{-67.899}\\

    AFlow        & 39.44 & 35.54 & 34.03 & 37.38 & 36.60 & 32.462 &5.00 & 40.19 & 24.16 & 14.02 & 20.84 & 7.561\\
    \rowcolor{unifiedColor}
    \tagUnified  & 48.33 & \textbf{37.69} & 44.29 & \textbf{54.21} & \textbf{46.13}  \improve{+9.53} & 32.861 \improve{+0.399}& 14.44 & \textbf{49.97} & 41.82 & 38.97 & \textbf{36.30} \improve{+15.46}& 7.734 \improve{+0.173}\\

    ScoreFlow     &32.22 & 31.00 & 36.10 & 18.69 & 29.50 & 36.908 &6.67 & 41.57 & 30.52 & 9.35 & 22.03 & 5.914\\
    \rowcolor{unifiedColor}
    \tagUnified  &39.44 & 32.37 & 44.55 & 50.47 & 41.71 \improve{+12.21} & 29.071 \improve{-7.837}& 7.78 & 43.37 & 34.03 & \textbf{40.19} & 31.34  \improve{+9.31}& 5.969 \improve{+0.055}\\

    MAS$^2$      &42.22 & 33.07 & 34.94 & 17.76 & 32.00 & 24.174 &3.89 & 42.41 & 32.34 & 15.89 & 23.63 & 3.368\\
    \rowcolor{unifiedColor}
    \tagUnified  & \textbf{48.89} & 35.09 & 46.25 & 49.06 & 44.82 \improve{+12.82} & 14.819 \improve{-9.355} &4.44 & 46.64 & \textbf{42.73} & 36.89 & 32.68 \improve{+9.05} & 2.858 \improve{-0.510}\\
    
    \midrule


    \textbf{Method} & \multicolumn{6}{c|}{\textbf{Qwen3-Next-80B-A3B-Instruct}} & \multicolumn{6}{c}{\textbf{DeepSeek-V3.2}} \\
    \midrule
    
    Vanilla       & 2.22 & 20.07 & 27.66 & 23.36 & 18.33 & 0.176 & 8.33 & 23.51 & 31.69 & 26.17 & 22.43 & 0.244 \\
    
    Manual MAS         & 8.89 & 27.27 & 35.97 & 32.71 & 26.21 & 5.867& 21.67 & 36.54 & 39.48 & 42.06 & 34.94 & 8.149 \\
    MetaAgent   & 1.67 & 21.22 & 29.74 & 10.28 & 15.73 & 2.148 & 0.56 & 23.84 & 32.47 & 28.97 & 21.46 & 2.586 \\
    EvoAgent   & 1.67 & 14.03 & 38.04 & 23.36 & 19.28 & 6.711 & 5.00 & 30.76 & 40.37 & 33.64 & 27.44 & 8.358 \\
    AOrchestra   &1.11 & 20.95 & 34.81 & 35.51 & 23.10 & 1.613 & 4.44 & 26.94 & 36.88 & 25.23 & 23.37 & 2.198 \\
    \midrule
    
    MAS-Zero   & 3.13 & 15.25 & 34.06 & 26.56 & 19.75 & 29.911 & 9.38 & 25.26 & 36.41 & 31.25 & 25.58 & 36.804 \\
    \rowcolor{unifiedColor}
    \tagUnified  &9.40 & 20.90 & 40.16 & 33.90 & 26.09 \improve{+6.34} & 18.939 \improve{-10.972} & \textbf{23.44} & 33.64 & 47.58 & 39.06 & 35.93 \improve{+10.35} & 18.622 \improve{-18.182} \\

    AFlow     & 3.33 & 25.81 & 30.26 & 24.30 & 20.93 & 1.678 & 12.22 & 25.56 & 38.44 & 36.45 & 28.17 & 7.773 \\
    \rowcolor{unifiedColor}
    \tagUnified  &5.56 & 32.46 & 37.40 & \textbf{56.08} & 32.88 \improve{+11.95} & 1.665 \improve{-0.013} & 22.22 & \textbf{39.51} & 50.26 & \textbf{46.73} & \textbf{39.68} \improve{+11.51} & 2.593 \improve{-5.180} \\

    ScoreFlow     & 5.00 & 24.31 & 35.71 & 32.64 & 24.42 & 4.585 & 15.00 & 25.85 & 38.31 & 40.19 & 29.84 & 7.361 \\
    \rowcolor{unifiedColor}
    \tagUnified  &10.56 & 31.55 & 39.87 & 53.27 & \textbf{33.81} \improve{+9.39} & 3.849 \improve{-0.736} & 18.89 & 34.60 & \textbf{54.81} & 45.79 & 38.52 \improve{+8.68} & 5.924 \improve{-1.437}\\

    MAS$^2$      & 5.00 & 27.10 & 32.47 & 30.84 & 23.85 & 2.000 & 10.00 & 26.81 & 37.27 & 27.10 & 25.30 & 1.650 \\
    \rowcolor{unifiedColor}
    \tagUnified  &\textbf{11.11} & \textbf{32.55} & \textbf{43.81} & 42.99 & 32.62 \improve{+8.77} & 1.008 \improve{-0.992} & 17.22 & 32.41 & 52.86 & 38.32 & 35.20 \improve{+9.90} & 1.338 \improve{-0.312} \\
    
    \bottomrule
    \end{tabular}
    }
    
    \caption{Quantification comparison of Unified-MAS and baselines on four benchmarks. Rows highlighted in \colorbox{unifiedColor}{blue} indicate methods with domain-specific nodes generated by Unified-MAS. \textbf{TP}: TravelPlanner, \textbf{HB}: HealthBench, \textbf{J1}: J1Bench, \textbf{DF}: DeepFund. \textbf{Avg.} reports average performance and cost. \textbf{Bold} denotes the best result.}
    \label{tab:main_results}
\end{table*}

\paragraph{Test Models.}
We deploy the \emph{same} LLM for every component within the final Automatic-MAS setups for fair comparison. Our evaluation spans four different models, including two closed-source models, Gemini-3-Flash~\cite{team2023gemini} and GPT-5-Mini~\cite{singh2025openai}, and two open-source models, Qwen3-Next-80B-A3B-Instruct~\cite{qwen3technicalreport} and DeepSeek-V3.2~\cite{liu2025deepseek}. 
Key configurations are documented in Appendix~\ref{appendix:experimental details}, and prompts for Unified-MAS are listed in Appendix~\ref{appendix:prompt details}.

\section{Results and Analysis}
\label{sec:results}

\subsection{Main Results}
\label{sec:main results}

\textbf{The Domain Barrier: Manual vs. Automatic-MAS. }
Table~\ref{tab:main_results} shows that task-specific Manual MAS consistently outperforms Automatic-MAS baselines across nearly all settings. For example, with Gemini-3-Flash, Manual MAS achieves an average score of 40.99, exceeding all Automatic-MAS baselines. This gap highlights the importance of domain expertise in complex tasks. Even with dynamic node generation, general-purpose orchestrators struggle to discover effective topologies without incorporating specialized knowledge.

\textbf{Trap of Dynamic Node Generation.}
Methods attempting dynamic node generation (i.e., MetaAgent, EvoAgent, AOrchestra) exhibit flashes of potential but suffer from severe systemic instability. For example, while EvoAgent marginally surpasses Manual MAS on J1Bench (e.g., 41.82 vs. 40.00 with Gemini-3-Flash), these dynamic methods fail catastrophically on TravelPlanner, often performing worse than the Vanilla baseline. 

\begin{figure}[t]
    \centering
    \includegraphics[width=\linewidth]{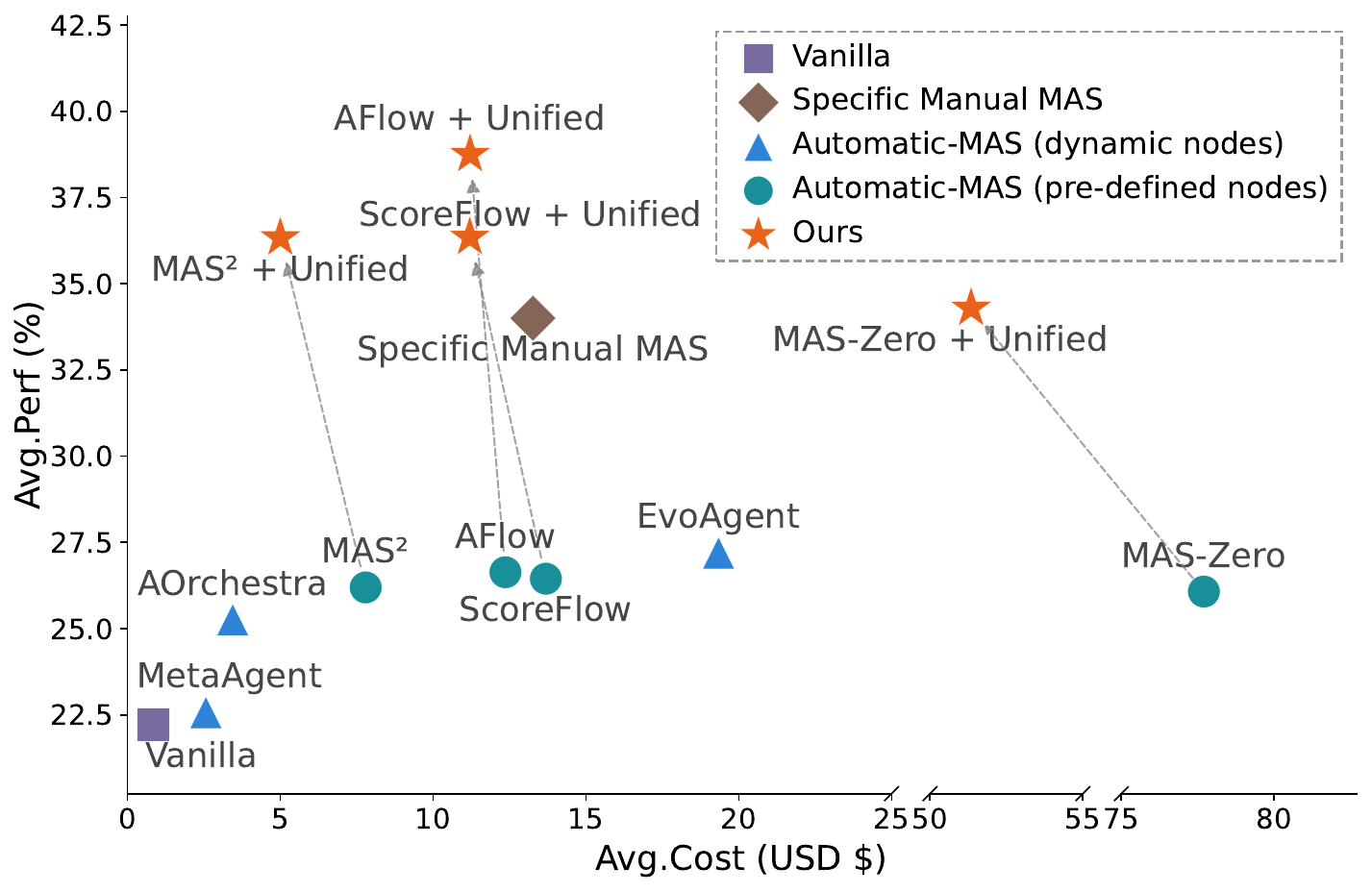} 
    \caption{Performance-cost trade-off averaged across four LLMs. Gray arrows illustrate Unified-MAS elevating baselines to higher performance at reduced costs.}
    \label{fig:cost analysis}
\end{figure}

\textbf{Unified-MAS Improves Performance and Efficiency.}
As shown in Table~\ref{tab:main_results}, integrating the domain-specific node set generated by Unified-MAS substantially improves the performance of predefined Automatic-MAS while universally reducing costs. In terms of average performance, incorporating domain-specific nodes yields consistent improvements across all settings, with gains ranging from 6.0\% (MAS-Zero with Qwen3-Next-80B-A3B-Instruct) to 14.2\% (AFlow with GPT-5-Mini).
Figure~\ref{fig:cost analysis} further demonstrates that methods enhanced by Unified-MAS consistently achieve a superior performance–cost trade-off compared to both manual and unenhanced automatic baselines\footnote{\arrmay{The difference between online inference cost and offline generation cost is detailed in Appendix~\ref{appendix:cost analysis}.}}. By replacing inefficient general nodes with optimized domain-specific nodes, Unified-MAS enables the system to solve complex problems with fewer and more effective steps.  These results confirm that Unified-MAS successfully bridges the gap, combining the reliability of expert nodes with the scalability of automated design.

\begin{table}[t]
    \centering
    
    \resizebox{0.48\textwidth}{!}{%

    \begin{tabular}{@{}llcccc@{}}
    \toprule
    \multirow{2.5}{*}{\textbf{Method}} & \multirow{2.5}{*}{\textbf{Designer}} & \multicolumn{2}{c}{\textbf{GPT-5-Mini}} & \multicolumn{2}{c}{\textbf{DeepSeek-V3.2}} \\
        \cmidrule(lr){3-4} \cmidrule(lr){5-6}
         & & Perf $\uparrow$ & Cost $\downarrow$ & Perf $\uparrow$ & Cost $\downarrow$ \\ \midrule
    AFlow & - & 20.84 & 7.561 & 28.17 & 7.773 \\
    \multirow{3}{*}{\tagUnified} & Gemini-3-Pro & \textbf{36.30} & 7.734 & \textbf{39.68} & 2.593 \\
     & Gemini-3-Flash & 33.46 & \textbf{4.421} & 37.14 & 2.221 \\
     & GPT-5-Mini & 35.84 & 7.093 & 35.49 & \textbf{2.125} \\ \midrule
    MAS$^2$ & - & 23.63 & 3.368 & 25.30 & 1.650 \\
    \multirow{3}{*}{\tagUnified} & Gemini-3-Pro & \textbf{32.68} & \textbf{2.858} & \textbf{35.20} & \textbf{1.338} \\
     & Gemini-3-Flash & 30.04 & 4.713 & 32.03 & 1.615 \\
     & GPT-5-Mini & 30.57 & 5.987 & 32.55 & 2.204 \\
        
    \bottomrule
    \end{tabular}
    }
    
    \caption{Robustness across different Designer LLMs.}
    \label{tab:different llms}
\end{table}

\textbf{Ablation Study.} 
As shown in Figure~\ref{fig:different epoch}, the results at epoch 0 and 10 correspond to the ablation variant \textbf{``Stage 1 only''} and \textbf{``Stage 1 + stage 2''}. First, the latter markedly outperforms the former. Second, even ``Stage 1 only'' already outperforms existing automatic baselines. Therefore, both stages contribute meaningfully to the overall performance.
\emnlpcr{
On one hand, in real-world scenarios where labeled validation data is unavailable, Unified-MAS can still be directly and effectively deployed using only the first stage. 
On the other hand, Stage-1 nodes can also be further improved by label-free self-evolution techniques, such as self-consistency, reflection, LLM-judge feedback, or online execution feedback, without being strictly limited to ground-truth validation labels.
}
More ablation studies can be found in Appendix~\ref{appendix:further ablation}.

\subsection{Further Analysis}

\subsubsection{Robustness to Designer Choices}
\label{sec:different designers}
Table~\ref{tab:different llms} reveals that Unified-MAS universally elevates baseline performance across all three Designers, demonstrating that Unified-MAS is highly robust to the choice of the ``Designer LLM''. Interestingly, we observe an architectural divergence based on the LLM's inherent preferences (Appendix~\ref{appendix:case study}). Gemini models tend to synthesize concise, macro-level workflows (5-6 nodes), whereas GPT-5-Mini prefers micro-level granularity (about 10 nodes by decomposing complex nodes further). Despite these distinct topological preferences, Unified-MAS is not bottlenecked by any single LLM, consistently driving substantial performance gains.

\subsubsection{Generalizability to General Domains}
\label{sec:on general domain}
\begin{table}[t]
    \centering
    
    \resizebox{0.48\textwidth}{!}{%

    \begin{tabular}{@{}lcccc@{}}
    \toprule
    \multirow{2.5}{*}{\textbf{Method}} & \multicolumn{2}{c}{\textbf{GPT-5-Mini}} & \multicolumn{2}{c}{\textbf{DeepSeek-V3.2}} \\
        \cmidrule(lr){2-3} \cmidrule(lr){4-5}
         & Perf $\uparrow$ & Cost $\downarrow$ & Perf $\uparrow$ & Cost $\downarrow$ \\ \midrule
        Vanilla   & 55.10  & 0.234 & 42.86 & 0.063 \\
        MAS-Zero   & 57.14 & 30.546 & 48.98 & 12.162 \\
        \tagUnified  & 59.18 \improve{+2.04} & 19.351 \improve{-11.195} & 53.06 \improve{+4.08} & 8.899 \improve{-3.263} \\

        AFlow        & 59.18 & 1.736 & 48.98 & 0.472 \\
        \tagUnified  & 67.35 \improve{+8.17} & 3.209 \improve{+1.473} & 55.10 \improve{+6.12} & 0.718 \improve{+0.246} \\

        ScoreFlow     & 57.14 & 0.701 & 44.90 & 0.305 \\
        \tagUnified  & 61.22 \improve{+4.08} & 0.854 \improve{+0.153} & 57.14 \improve{+12.24} & 0.462 \improve{+0.157} \\

        MAS$^2$      & 63.27 & 1.133 & 51.02 & 0.434 \\
        \tagUnified  & 67.35 \improve{+4.08} & 1.040 \improve{-0.093} & 66.67 \improve{+15.65} & 0.884 \improve{+0.450} \\
        
    \bottomrule
    \end{tabular}
    }
    \caption{Results of General Automatic-MAS with/without Unified-MAS on AIME24\&25.}
    \label{tab:aime}
\end{table}

While our main evaluation focuses on specialized domains, Table~\ref{tab:aime} extends the analysis to general domains (mathematical reasoning) using AIME 2024 and 2025~\cite{aime}. Integrating Unified-MAS consistently improves performance across all baselines for both GPT-5-Mini and DeepSeek-V3.2. Although the gains are more modest than the substantial improvements observed in knowledge-intensive tasks, the results prove that our framework can successfully synthesize reasonable, fine-grained mathematical nodes (see Appendix~\ref{appendix:case study}), demonstrating broad applicability even in conventional reasoning tasks. \arrmay{We further discuss the scalability and adaptation problem of Unified-MAS in Appendix~\ref{appendix:discussion}.}

\subsubsection{Successful Pattern}
\label{sec:successful pattern}
To understand this performance leap, we qualitatively compare the nodes generated by Unified-MAS against those from dynamic Automatic-MAS on J1Bench (Appendix~\ref{appendix:case study}). 
Dynamic methods like EvoAgent resort to a lazy ensemble approach, generating superficial nodes like ``Expert1'' and ``Expert2'' without true domain grounding. In sharp contrast, Unified-MAS synthesizes a highly structured, expert-level judicial pipeline. It explicitly divides reasoning into professional stages: ``Legal\_Element\_Extractor'', ``Liability\_Reasoning'', and so on. As detailed in Appendix~\ref{appendix:case study}, compared to the blind prompt-level voting of original AFlow, the Unified-MAS-enhanced workflow ensures that every stage is traceable and legally grounded.

\subsubsection{The Optimization Dynamics}
\label{sec:optimization epoch}
\begin{figure}[t]
    \centering
    \includegraphics[width=\linewidth]{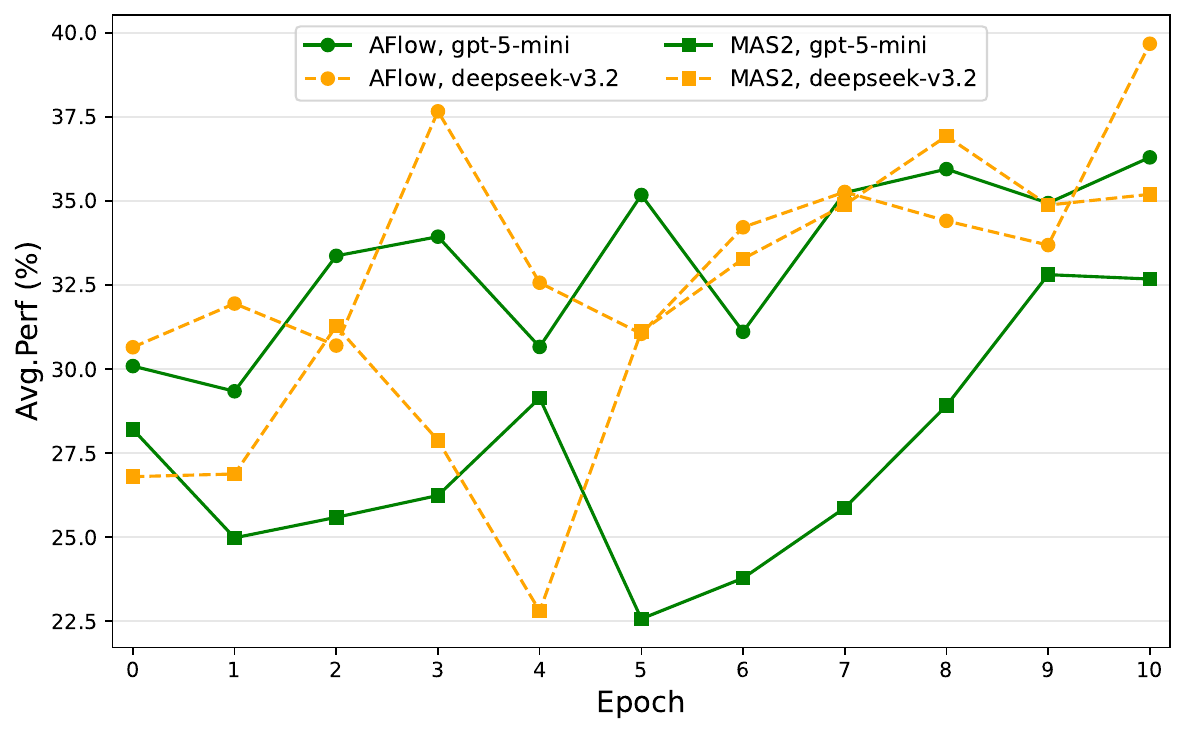} 
    \caption{Epoch-wise performance dynamics during node optimization using Gemini-3-Pro as the Designer.}
    \label{fig:different epoch}
\end{figure}

Our reward-based node optimization reveals an important learning dynamic. As shown in Figure~\ref{fig:different epoch}, the performance trajectory is non-monotonic. From our observation, during early epochs (0 to 5), the system repeatedly targets the most severe ``bottleneck node''. Updating this node temporarily disrupts established cross-node co-adaptations, causing short-term perturbation. However, once the bottleneck is sufficiently alleviated, the system shifts focus to other nodes. Consequently, performance rapidly recovers and converges to a sustained global optimum in the later epochs (6–10). These results indicate that our node optimization strategy effectively removes brittle internal logic while avoiding trapping the system in local optima.

\section{Conclusion}
\label{sec:conclusion}

In this work, we decouple granular node implementation from topology orchestration and propose Unified-MAS, which automatically synthesizes domain-specific nodes through external knowledge retrieval and iteratively refines them via a perplexity-guided reward. Extensive experiments demonstrate that integrating our generated nodes into existing Automatic-MAS approaches universally enhances overall performance, yielding improvements of up to 14.2\% while simultaneously reducing costs. Further analysis highlights the robustness of Unified-MAS across different Designer LLMs, demonstrates its generalizability to general domains, and confirms the critical role of the reward-based optimization stage. Moving forward, Unified-MAS can be broadly applied to virtually any specific domain to generate highly professional nodes, seamlessly bridging the gap between general Automatic-MAS and deep domain expertise for future scalable real-world applications.

\section*{Limitations}
While Unified-MAS demonstrates significant efficacy, we acknowledge certain limitations that present exciting avenues for future research. Primarily, our current framework operates as an offline node-preparation phase, which restricts its immediate applicability in highly dynamic or extremely time-sensitive environments that necessitate real-time, on-the-fly node generation and adaptation. 
To transition towards fully online, adaptive synthesis, future work should proceed in two main directions. On one hand, future work should focus on streamlining the generation pipeline, allowing the framework to rapidly create and adapt nodes directly. On the other hand, future systems could learn directly from live feedback, quickly adjusting nodes instead of relying on a long offline evaluation.

\section*{Acknowledgements}
This work is supported by funding from ModelBest.


\emnlpcr{
\section*{Ethical Considerations}
Unified-MAS reduces the effort required to construct domain-specific multi-agent systems, but generated nodes can inherit factual errors, biases, privacy risks, and correlated failures from their underlying language models. The multi-agent system may produce misinformation, especially when deployed in real-world high-stakes domains such as healthcare, law, finance, or education. We recommend treating it as an assistant system rather than an independent expert, minimizing sensitive data exposure, using least-privilege tool access and human review for consequential outputs. 
}

\bibliography{custom}

@inproceedings{he2025self,
  title={Self-correction is more than refinement: A learning framework for visual and language reasoning tasks},
  author={He, Jiayi and Lin, Hehai and Wang, Qingyun and Fung, Yi R and Ji, Heng},
  booktitle={Findings of the Association for Computational Linguistics: ACL 2025},
  pages={6405--6421},
  year={2025}
}

@article{ferrag2025llm,
  title={From llm reasoning to autonomous ai agents: A comprehensive review},
  author={Ferrag, Mohamed Amine and Tihanyi, Norbert and Debbah, Merouane},
  journal={arXiv preprint arXiv:2504.19678},
  year={2025}
}

@article{ke2025survey,
  title={A survey of frontiers in llm reasoning: Inference scaling, learning to reason, and agentic systems},
  author={Ke, Zixuan and Jiao, Fangkai and Ming, Yifei and Nguyen, Xuan-Phi and Xu, Austin and Long, Do Xuan and Li, Minzhi and Qin, Chengwei and Wang, Peifeng and Savarese, Silvio and others},
  journal={arXiv preprint arXiv:2504.09037},
  year={2025}
}

@inproceedings{du2024improving,
  title={Improving factuality and reasoning in language models through multiagent debate},
  author={Du, Yilun and Li, Shuang and Torralba, Antonio and Tenenbaum, Joshua B and Mordatch, Igor},
  booktitle={Forty-first international conference on machine learning},
  year={2024}
}

@article{lin2025interactive,
  title={Interactive learning for LLM reasoning},
  author={Lin, Hehai and Cao, Shilei and Wang, Sudong and Wu, Haotian and Li, Minzhi and Yang, Linyi and Zheng, Juepeng and Qin, Chengwei},
  journal={arXiv preprint arXiv:2509.26306},
  year={2025}
}

@article{wang2022self,
  title={Self-consistency improves chain of thought reasoning in language models},
  author={Wang, Xuezhi and Wei, Jason and Schuurmans, Dale and Le, Quoc and Chi, Ed and Narang, Sharan and Chowdhery, Aakanksha and Zhou, Denny},
  journal={arXiv preprint arXiv:2203.11171},
  year={2022}
}

@article{shinn2023reflexion,
  title={Reflexion: Language agents with verbal reinforcement learning},
  author={Shinn, Noah and Cassano, Federico and Gopinath, Ashwin and Narasimhan, Karthik and Yao, Shunyu},
  journal={Advances in neural information processing systems},
  volume={36},
  pages={8634--8652},
  year={2023}
}

@article{ke2025mas,
  title={Mas-zero: Designing multi-agent systems with zero supervision},
  author={Ke, Zixuan and Xu, Austin and Ming, Yifei and Nguyen, Xuan-Phi and Chin, Ryan and Xiong, Caiming and Joty, Shafiq},
  journal={arXiv preprint arXiv:2505.14996},
  year={2025}
}

@article{ye2025mas,
  title={Mas-gpt: Training llms to build llm-based multi-agent systems},
  author={Ye, Rui and Tang, Shuo and Ge, Rui and Du, Yaxin and Yin, Zhenfei and Chen, Siheng and Shao, Jing},
  journal={arXiv preprint arXiv:2503.03686},
  year={2025}
}

@article{tran2025multi,
  title={Multi-agent collaboration mechanisms: A survey of llms},
  author={Tran, Khanh-Tung and Dao, Dung and Nguyen, Minh-Duong and Pham, Quoc-Viet and O'Sullivan, Barry and Nguyen, Hoang D},
  journal={arXiv preprint arXiv:2501.06322},
  year={2025}
}

@article{hu2024automated,
  title={Automated design of agentic systems},
  author={Hu, Shengran and Lu, Cong and Clune, Jeff},
  journal={arXiv preprint arXiv:2408.08435},
  year={2024}
}

@article{ke2026mas,
  title={MAS-Orchestra: Understanding and Improving Multi-Agent Reasoning Through Holistic Orchestration and Controlled Benchmarks},
  author={Ke, Zixuan and Ming, Yifei and Xu, Austin and Chin, Ryan and Nguyen, Xuan-Phi and Jwalapuram, Prathyusha and Yavuz, Semih and Xiong, Caiming and Joty, Shafiq},
  journal={arXiv preprint arXiv:2601.14652},
  year={2026}
}

@article{xu2025staf,
  title={Staf-llm: A scalable and task-adaptive fine-tuning framework for large language models in medical domain},
  author={Xu, Tianhan and Chen, Ling and Hu, Zhe and Li, Bin},
  journal={Expert Systems with Applications},
  volume={281},
  pages={127582},
  year={2025},
  publisher={Elsevier}
}

@article{li2024survey,
  title={A survey on LLM-based multi-agent systems: workflow, infrastructure, and challenges},
  author={Li, Xinyi and Wang, Sai and Zeng, Siqi and Wu, Yu and Yang, Yi},
  journal={Vicinagearth},
  volume={1},
  number={1},
  pages={9},
  year={2024},
  publisher={Springer}
}

@article{ruan2026aorchestra,
  title={AOrchestra: Automating Sub-Agent Creation for Agentic Orchestration},
  author={Ruan, Jianhao and Xu, Zhihao and Peng, Yiran and Ren, Fashen and Yu, Zhaoyang and Liang, Xinbing and Xiang, Jinyu and Liu, Bang and Wu, Chenglin and Luo, Yuyu and others},
  journal={arXiv preprint arXiv:2602.03786},
  year={2026}
}

@article{zhang2025metaagent,
  title={Metaagent: Automatically constructing multi-agent systems based on finite state machines},
  author={Zhang, Yaolun and Liu, Xiaogeng and Xiao, Chaowei},
  journal={arXiv preprint arXiv:2507.22606},
  year={2025}
}

@article{xie2024travelplanner,
  title={Travelplanner: A benchmark for real-world planning with language agents},
  author={Xie, Jian and Zhang, Kai and Chen, Jiangjie and Zhu, Tinghui and Lou, Renze and Tian, Yuandong and Xiao, Yanghua and Su, Yu},
  journal={arXiv preprint arXiv:2402.01622},
  year={2024}
}

@article{arora2025healthbench,
  title={Healthbench: Evaluating large language models towards improved human health},
  author={Arora, Rahul K and Wei, Jason and Hicks, Rebecca Soskin and Bowman, Preston and Qui{\~n}onero-Candela, Joaquin and Tsimpourlas, Foivos and Sharman, Michael and Shah, Meghan and Vallone, Andrea and Beutel, Alex and others},
  journal={arXiv preprint arXiv:2505.08775},
  year={2025}
}

@article{jia2025ready,
  title={Ready jurist one: Benchmarking language agents for legal intelligence in dynamic environments},
  author={Jia, Zheng and Yue, Shengbin and Chen, Wei and Wang, Siyuan and Liu, Yidong and Li, Zejun and Song, Yun and Wei, Zhongyu},
  journal={arXiv preprint arXiv:2507.04037},
  year={2025}
}

@article{li2025time,
  title={Time Travel is Cheating: Going Live with DeepFund for Real-Time Fund Investment Benchmarking},
  author={Li, Changlun and Shi, Yao and Wang, Chen and Duan, Qiqi and Ruan, Runke and Huang, Weijie and Long, Haonan and Huang, Lijun and Tang, Nan and Luo, Yuyu},
  journal={arXiv preprint arXiv:2505.11065},
  year={2025}
}

@article{zhang2024aflow,
  title={Aflow: Automating agentic workflow generation},
  author={Zhang, Jiayi and Xiang, Jinyu and Yu, Zhaoyang and Teng, Fengwei and Chen, Xionghui and Chen, Jiaqi and Zhuge, Mingchen and Cheng, Xin and Hong, Sirui and Wang, Jinlin and others},
  journal={arXiv preprint arXiv:2410.10762},
  year={2024}
}

@article{wang2025scoreflow,
  title={Scoreflow: Mastering llm agent workflows via score-based preference optimization},
  author={Wang, Yinjie and Yang, Ling and Li, Guohao and Wang, Mengdi and Aragam, Bryon},
  journal={arXiv preprint arXiv:2502.04306},
  year={2025}
}

@article{wang2025mas,
  title={MAS $^2$: Self-Generative, Self-Configuring, Self-Rectifying Multi-Agent Systems},
  author={Wang, Kun and Zhang, Guibin and Ye, ManKit and Deng, Xinyu and Wang, Dongxia and Hu, Xiaobin and Guo, Jinyang and Liu, Yang and Guo, Yufei},
  journal={arXiv preprint arXiv:2509.24323},
  year={2025}
}

@article{madaan2023self,
  title={Self-refine: Iterative refinement with self-feedback},
  author={Madaan, Aman and Tandon, Niket and Gupta, Prakhar and Hallinan, Skyler and Gao, Luyu and Wiegreffe, Sarah and Alon, Uri and Dziri, Nouha and Prabhumoye, Shrimai and Yang, Yiming and others},
  journal={Advances in neural information processing systems},
  volume={36},
  pages={46534--46594},
  year={2023}
}

@article{chen2024survey,
  title={A survey on llm-based multi-agent system: Recent advances and new frontiers in application},
  author={Chen, Shuaihang and Liu, Yuanxing and Han, Wei and Zhang, Weinan and Liu, Ting},
  journal={arXiv preprint arXiv:2412.17481},
  year={2024}
}

@article{liu2023dynamic,
  title={Dynamic llm-agent network: An llm-agent collaboration framework with agent team optimization},
  author={Liu, Zijun and Zhang, Yanzhe and Li, Peng and Liu, Yang and Yang, Diyi},
  journal={arXiv preprint arXiv:2310.02170},
  year={2023}
}

@inproceedings{yuan2025evoagent,
  title={Evoagent: Towards automatic multi-agent generation via evolutionary algorithms},
  author={Yuan, Siyu and Song, Kaitao and Chen, Jiangjie and Tan, Xu and Li, Dongsheng and Yang, Deqing},
  booktitle={Proceedings of the 2025 Conference of the Nations of the Americas Chapter of the Association for Computational Linguistics: Human Language Technologies (Volume 1: Long Papers)},
  pages={6192--6217},
  year={2025}
}

@article{ji2023survey,
  title={Survey of hallucination in natural language generation},
  author={Ji, Ziwei and Lee, Nayeon and Frieske, Rita and Yu, Tiezheng and Su, Dan and Xu, Yan and Ishii, Etsuko and Bang, Ye Jin and Madotto, Andrea and Fung, Pascale},
  journal={ACM computing surveys},
  volume={55},
  number={12},
  pages={1--38},
  year={2023},
  publisher={ACM New York, NY}
}

@inproceedings{hong2023metagpt,
  title={MetaGPT: Meta programming for a multi-agent collaborative framework},
  author={Hong, Sirui and Zhuge, Mingchen and Chen, Jonathan and Zheng, Xiawu and Cheng, Yuheng and Wang, Jinlin and Zhang, Ceyao and Wang, Zili and Yau, Steven Ka Shing and Lin, Zijuan and others},
  booktitle={The twelfth international conference on learning representations},
  year={2023}
}

@article{huang2026ama,
  title={AMA: Adaptive Memory via Multi-Agent Collaboration},
  author={Huang, Weiquan and Wang, Zixuan and Lin, Hehai and Wang, Sudong and Xu, Bo and Li, Qian and Zhu, Beier and Yang, Linyi and Qin, Chengwei},
  journal={arXiv preprint arXiv:2601.20352},
  year={2026}
}

@article{liu2025rectifying,
  title={Rectifying LLM Thought from Lens of Optimization},
  author={Liu, Junnan and Liu, Hongwei and Zhang, Songyang and Chen, Kai},
  journal={arXiv preprint arXiv:2512.01925},
  year={2025}
}

@article{kendall1938new,
  title={A new measure of rank correlation},
  author={Kendall, Maurice G},
  journal={Biometrika},
  volume={30},
  number={1-2},
  pages={81--93},
  year={1938},
  publisher={Oxford University Press}
}

@inproceedings{zhang2025planning,
  title={Planning with multi-constraints via collaborative language agents},
  author={Zhang, Cong and Goh, Xin Deik and Li, Dexun and Zhang, Hao and Liu, Yong},
  booktitle={Proceedings of the 31st International Conference on Computational Linguistics},
  pages={10054--10082},
  year={2025}
}

@article{wu2025furina,
  title={FURINA: A Fully Customizable Role-Playing Benchmark via Scalable Multi-Agent Collaboration Pipeline},
  author={Wu, Haotian and Jiang, Shufan and Chen, Mingyu and Feng, Yiyang and Lin, Hehai and Zou, Heqing and Shu, Yao and Qin, Chengwei},
  journal={arXiv preprint arXiv:2510.06800},
  year={2025}
}

@article{zhao2026training,
  title={Training Multi-Turn Search Agent via Contrastive Dynamic Branch Sampling},
  author={Zhao, Yubao and Huang, Weiquan and Wang, Sudong and Zhao, Ruochen and Chen, Chen and Shu, Yao and Qin, Chengwei},
  journal={arXiv preprint arXiv:2602.03719},
  year={2026}
}

@article{chen2025enhancing,
  title={Enhancing diagnostic capability with multi-agents conversational large language models},
  author={Chen, Xi and Yi, Huahui and You, Mingke and Liu, WeiZhi and Wang, Li and Li, Hairui and Zhang, Xue and Guo, Yingman and Fan, Lei and Chen, Gang and others},
  journal={NPJ digital medicine},
  volume={8},
  number={1},
  pages={159},
  year={2025},
  publisher={Nature Publishing Group UK London}
}

@misc{aime,
  title={AIME problems and solutions.},
  author={MAA-Committees},
  url={https://artofproblemsolving.com/wiki/index.php/AIME_Problems_and_Solutions},
  year={2025}
}

@article{liu2025deepseek,
  title={Deepseek-v3. 2: Pushing the frontier of open large language models},
  author={Liu, Aixin and Mei, Aoxue and Lin, Bangcai and Xue, Bing and Wang, Bingxuan and Xu, Bingzheng and Wu, Bochao and Zhang, Bowei and Lin, Chaofan and Dong, Chen and others},
  journal={arXiv preprint arXiv:2512.02556},
  year={2025}
}

@misc{qwen3technicalreport,
      title={Qwen3 Technical Report}, 
      author={Qwen Team},
      year={2025},
      eprint={2505.09388},
      archivePrefix={arXiv},
      primaryClass={cs.CL},
      url={https://arxiv.org/abs/2505.09388}, 
}

@article{singh2025openai,
  title={Openai gpt-5 system card},
  author={Singh, Aaditya and Fry, Adam and Perelman, Adam and Tart, Adam and Ganesh, Adi and El-Kishky, Ahmed and McLaughlin, Aidan and Low, Aiden and Ostrow, AJ and Ananthram, Akhila and others},
  journal={arXiv preprint arXiv:2601.03267},
  year={2025}
}

@article{team2023gemini,
  title={Gemini: a family of highly capable multimodal models},
  author={Team, Gemini and Anil, Rohan and Borgeaud, Sebastian and Alayrac, Jean-Baptiste and Yu, Jiahui and Soricut, Radu and Schalkwyk, Johan and Dai, Andrew M and Hauth, Anja and Millican, Katie and others},
  journal={arXiv preprint arXiv:2312.11805},
  year={2023}
}

@article{huang2025survey,
  title={A survey on hallucination in large language models: Principles, taxonomy, challenges, and open questions},
  author={Huang, Lei and Yu, Weijiang and Ma, Weitao and Zhong, Weihong and Feng, Zhangyin and Wang, Haotian and Chen, Qianglong and Peng, Weihua and Feng, Xiaocheng and Qin, Bing and others},
  journal={ACM Transactions on Information Systems},
  volume={43},
  number={2},
  pages={1--55},
  year={2025},
  publisher={ACM New York, NY}
}

@article{dubey2024llama,
  title={The llama 3 herd of models},
  author={Dubey, Abhimanyu and Jauhri, Abhinav and Pandey, Abhinav and Kadian, Abhishek and Al-Dahle, Ahmad and Letman, Aiesha and Mathur, Akhil and Schelten, Alan and Yang, Amy and Fan, Angela and others},
  journal={arXiv e-prints},
  pages={arXiv--2407},
  year={2024}
}

@article{jwalapuram2026illusion,
  title={The Illusion of Multi-Agent Advantage},
  author={Jwalapuram, Prathyusha and Lin, Hehai and Li, Chuyuan and Jiao, Fangkai and Wang, Sudong and Ming, Yifei and Ke, Zixuan and Qin, Chengwei and Carenini, Giuseppe and Joty, Shafiq},
  journal={arXiv preprint arXiv:2606.13003},
  year={2026}
}

@article{lin2026skill,
  title={Skill-MAS: Evolving Meta-Skill for Automatic Multi-Agent Systems},
  author={Lin, Hehai and Yang, Qi and Qin, Chengwei},
  journal={arXiv preprint arXiv:2606.18837},
  year={2026}
}

@misc{yu2026fidesfaithfulinferencedeep,
      title={FIDES: Faithful Inference via Deep Evidence Signals for Retrieval-Memory Conflict in RAG}, 
      author={Zhe Yu and Wenpeng Xing and Tiancheng Zhao and Mohan Li and Changting Lin and Meng Han},
      year={2026},
      eprint={2606.05644},
      archivePrefix={arXiv},
      primaryClass={cs.AI},
      url={https://arxiv.org/abs/2606.05644}, 
}

@misc{yu2026compositioncollapsestablefactual,
      title={Composition Collapse: Stable Factual Knowledge Does Not Imply Compositional Reasoning}, 
      author={Zhe Yu and Wenpeng Xing and Yunzhao Wei and Jie Chen and Hongzhi Wang and Xuyang Teng and Meng Han},
      year={2026},
      eprint={2605.26789},
      archivePrefix={arXiv},
      primaryClass={cs.AI},
      url={https://arxiv.org/abs/2605.26789}, 
}

\appendix

\section{Description of Appendix}
\label{appendix:description of appendix}

The appendix provides extended methodological details and comprehensive experimental data to further support the findings presented in the main manuscript. \textbf{Appendix~\ref{appendix:pseudocode}} presents the detailed pseudocode illustrating the algorithmic workflow of the proposed two-stage Unified-MAS. \textbf{Appendix~\ref{appendix:statistics of benchmarks}} provides exhaustive statistics and descriptive summaries of the diverse evaluation benchmarks, detailing dataset splitting protocols and the specific characteristics of each domain-specific task. \textbf{Appendix~\ref{appendix:experimental details}} delineates the experimental setup, including the baselines, the implementation details, cost analysis, retrieval leakage control, human expert evaluation, discussion, and further ablation studies. \textbf{Appendix~\ref{appendix:case study}} offers a qualitative case study that compares the node generation of Unified-MAS against existing Automatic-MAS. Finally, \textbf{Appendix~\ref{appendix:prompt details}} catalogs the comprehensive set of prompts utilized for Unified-MAS and our experiment.

\section{Pseudocode of Unified-MAS}
\label{appendix:pseudocode}
We provide the pseudocode of Unified-MAS here.

\begin{algorithm}[H] 
\caption{Unified-MAS}
\label{alg:unified_mas}
\begin{algorithmic}[1]
\REQUIRE Validation set $\mathcal{D}_{val}$, LLM $P_\theta$, Max epochs $K$, Balance factor $\alpha$, Sample size $N$
\ENSURE Domain-specific node set $\mathcal{V}_{domain}$

\textbf{\textit{Stage 1: Search-Based Node Generation}}
\STATE Sample $N$ examples from $\mathcal{D}_{val}$ to form $\mathcal{C}$
\STATE Extract keywords across 7 dimensions from $\mathcal{C}$
\STATE Synthesize search queries for 4 strategies
\STATE Retrieve external knowledge
\STATE Generate initial node set $\mathcal{V}_{init} = \{v_1, \dots, v_m\}$

\vspace{10pt}
\textbf{\textit{Stage 2: Reward-Based Node Optimization}}
\STATE $\mathcal{V}_{domain} \leftarrow \mathcal{V}_{init}$
\FOR{$k = 1$ to $K$}
    \STATE Initialize $R[v] \leftarrow \emptyset$ for all $v \in \mathcal{V}_{domain}$
    \FOR{each sample $(q, y) \in \mathcal{D}_{val}$}
        \STATE Initialize empty context $A_0 \leftarrow [h_0]$
        \STATE Compute baseline predictability: $\mathcal{J}_0 = -\log(\text{PPL}(y|q, A_0))$
        \FOR{$t = 1$ to $m$}
            \STATE Execute node $v_t$, obtain reasoning $h_t$
            \STATE Update accumulated context: $A_t \leftarrow [h_0, h_1, \dots, h_t]$
            \STATE Compute: $\mathcal{J}_t = -\log(\text{PPL}(y|q, A_t))$
            \STATE Calculate relative gain: $\delta_t = (\mathcal{J}_t - \mathcal{J}_0) / \mathcal{J}_0$
            \STATE Compute Improvement Score: $S_{i,t} = \tanh(\delta_t + 1)$
            \STATE Compute Consistency Score $\mathcal{S}_{c,t}$ using Eq. (6)
            \STATE Node Quality Score: $S_t = (1 - \alpha)S_{i,t} + \alpha\mathcal{S}_{c,t}$
            \IF{$t > 1$}
                \STATE Node reward: $r_t = S_t - S_{t-1}$
            \ELSE
                \STATE Node reward: $r_t = S_t$
            \ENDIF
            \STATE Append $r_t$ to $R[v_t]$
        \ENDFOR
    \ENDFOR
    \FOR{each node $v \in \mathcal{V}_{domain}$}
        \STATE Calculate average reward $\bar{r}(v)$ from $R[v]$
    \ENDFOR
    \STATE Identify $v^* = \arg\min_{v \in \mathcal{V}_{domain}} \bar{r}(v)$
    \STATE Retrieve samples where $v^*$ yielded the lowest reward and refine implementation
\ENDFOR
\RETURN $\mathcal{V}_{domain}$
\end{algorithmic}
\end{algorithm}

\section{Statistics of Benchmarks}
\label{appendix:statistics of benchmarks}

We split the entire dataset into the validation set and the test set because some Automatic-MAS needs the validation set to sample the best multi-agent system. For fair comparison, all the reported results are based on the test set. We randomly sample some examples from these datasets to build the validation and test set, which can be found in Table~\ref{tab:split of dataset}.

\textbf{TravelPlanner}~\cite{xie2024travelplanner}: This benchmark aims to evaluate the planning capabilities of language agents within complex, real-world travel scenarios. It features 1,225 meticulously curated user intents, and the evaluation focuses on an agent’s proficiency in multi-constraint reasoning and effective tool utilization, serving as a rigorous test for assessing how models navigate intricate planning tasks and integrate disparate information to achieve actionable objectives.

\textbf{HealthBench}~\cite{arora2025healthbench}: This benchmark is designed to evaluate the clinical proficiency and safety of AI agents in healthcare. Drawing upon the expertise of 262 practicing physicians across 60 countries, the dataset encompasses 5,000 authentic clinical dialogue scenarios ranging from acute emergencies to global health issues. Utilizing a physician-curated rubric, HealthBench moves beyond simple outcome metrics to rigorously assess models across critical dimensions, including clinical accuracy, communication quality, situational awareness, and safety, thereby ensuring robust performance in high-stakes medical applications.

\textbf{J1Bench}~\cite{jia2025ready}: This benchmark focuses on automated legal adjudication by simulating court proceedings. The input consists of 93 comprehensive cases, including formal complaints, defendant arguments, and evidentiary materials derived from actual judicial records. The agent is required to synthesize these conflicting testimonies and legal documents to produce a reasoned, final judicial judgment. Evaluation is based on the alignment of the agent's verdict with ground-truth, measuring the model's capacity to interpret legal arguments and arrive at legally sound conclusions.

\textbf{DeepFund}~\cite{li2025time}:  This benchmark evaluates the financial intelligence of agents in stock market decision-making. The input features a rich, time-sensitive dataset comprising corporate fundamental data, historical price trends, and real-time financial news streams. For a targeted list of stocks, the agent is tasked with outputting a categorical decision, specifically, ``Buy'', ``Sell'', or ``Hold''. The full dataset contains 139 cases to assess the agent's ability to effectively integrate heterogeneous information into actionable investment strategies.

\textbf{AIME24\&25}~\cite{aime}: This benchmark collection contains 57 questions and derives from the 2024 and 2025 editions of the American Invitational Mathematics Examination (AIME), comprising two distinct problem sets. Each set contains rigorously vetted mathematical questions characterized by high cognitive demand. The evaluative focus lies in probing advanced mathematical competencies, with particular emphasis on multi-faceted problem-solving strategies that require integration of complex conceptual frameworks.

\begin{table*}[htbp]
    \centering
    \begin{tabular}{@{}lccccc@{}}
    \toprule
    Split  & TravelPlanner & HealthBench & J1Bench & DeepFund & AIME24\&25 \\ \midrule
    Validation & 45 & 32 & 16 & 32 & 8 \\ 
    Test & 180 & 168 & 77 & 107 & 49 \\
    \bottomrule
    \end{tabular}%

    \caption{Data size for each split in each dataset.}
    \label{tab:split of dataset}
\end{table*}

\begin{table*}[htbp]
    \centering
    \begin{tabular}{@{}llcc@{}}
    \toprule
    Category & Hyperparameter  & Description & Value \\ \midrule
    \multirow{4}{*}{Unified-MAS}& $N$ & The number of samples used to build context buffer & 10 \\
    & Turn & The turn number of multi-turn search & 10 \\ 
    & $\alpha$ & The weight used to aggregate $\mathcal{S}_{i, t}$ and $\mathcal{S}_{c, t}$ & 0.6 \\
    & $K$ & The epoch number of node optimization stage & 10 \\
    \midrule
    \multirow{2}{*}{LLM calls}& temperature & The sampling temperature of calling LLM & 1.0 \\
    & max\_tokens & The maximum number of output tokens & 32768 \\
    \bottomrule
    \end{tabular}%

    \caption{The description and value of important hyperparameters.}
    \label{tab:hyper-parameters}
\end{table*}

\begin{table}[htbp]
    \centering
    \begin{tabular}{@{}l|cc@{}}
    \toprule
    Dataset & Generation & Optimization  \\ \midrule
    TravelPlanner & 10.780 & 4.001 \\
    HealthBench & 8.033 & 1.793 \\
    J1Bench & 11.093 & 1.737 \\
    DeepFund & 10.170 & 3.113 \\
    AIME24\&25 & 8.891 & 0.255 \\
    \bottomrule
    \end{tabular}%

    \caption{Cost (USD \$) of Unified-MAS using Gemini-3-Pro as the Designer.}
    \label{tab:unified-MAS cost}
\end{table}

\section{Experimental Details}
\label{appendix:experimental details}

\subsection{Specific Manual MAS Baselines}

\textbf{PMC}~\cite{zhang2025planning}:
PMC employs a hierarchical planning framework where a centralized planner decomposes complex tasks into sub-tasks, which are then executed by specialized agents with predefined roles. Incorporating a structured collaboration protocol, it ensures systematic problem-solving across multi-stage reasoning chains.

\textbf{Diagnosis-MAS}~\cite{chen2025enhancing}:
Diagnosis-MAS utilizes a multi-stage diagnostic workflow where agents engage in iterative feedback loops to identify and mitigate noise in reasoning processes. This approach systematically filters out erroneous information, thereby significantly enhancing the reliability of medical diagnosis.

\textbf{Court-MAS}~\cite{jia2025ready}:
Court-MAS adopts an adversarial interaction model inspired by judicial processes, where agents act as competing parties to present evidence and verify claims. A central judge-agent then adjudicates these contributions based on the simulated interaction.

\textbf{DeepFund-MAS}~\cite{li2025time}:
DeepFund-MAS implements a multi-agent architecture tailored for financial analysis, where agents are partitioned into functional units such as data acquisition, sentiment analysis, and risk assessment. The system allows agents to correlate disparate financial signals into coherent investment insights.

\subsection{Implementation Details}
\label{appendix:implementation details}

For cost considerations, we set AFlow's maximum number of iterations to 10 and run the validation set once each round. For all other baselines, we strictly follow the original settings. Table~\ref{tab:hyper-parameters} lists the important hyperparameters used in Unified-MAS. We set GPT-5-Mini with ``low'' reasoning effort, while leveraging the standard instruction versions of the other three LLMs. We use GPT-4o as the default LLM-judge following~\cite{ke2025mas}. 

\subsection{Cost Analysis}
\label{appendix:cost analysis}
\arrmay{
The node generation and iterative optimization stages are one-time, offline processes. Once the domain-specific node library is constructed, it is repeatedly reused by the Automatic-MAS orchestrator during inference. The inference cost is typically the bottleneck for real-world deployment. Therefore, the efficiency gains we emphasize in the main text and Table~\ref{tab:main_results} refer strictly to the online inference phase. 
}
Here we show the offline cost of Unified-MAS's two stages in Table~\ref{tab:unified-MAS cost}.
\arrmay{The offline node generation and optimization incur a one-time upfront cost. However, these optimized nodes will reduce the costs required during repeated online inference, leading to a much better long-term performance-cost trade-off.}

\subsection{Retrieval Leakage Control}
\label{appendix:leakage}
\arrmay{
We prevent data contamination during the search-based node generation stage in two ways. First, we instruct the LLM to search for domain logic and background knowledge, while explicitly forbidding it from searching for the input question itself. Second, we apply a hard rule-based filter to drop any retrieved documents containing the dataset names. After fetching all the contents, we conduct a rigorous LLM-based inspection on all retrieved documents (TravelPlanner: 289, HealthBench: 202, J1Bench: 266, DeepFund: 248, AIME: 224). The LLM evaluator is prompted to flag any document that: (1) contains the exact question text; (2) contains the direct final answer or official rubric; or (3) contains a paraphrased version preserving the same entities/numbers. The inspection confirms zero instances of contamination.
}

\emnlpcr{
\subsection{Human Expert Evaluation}
\label{appendix:human expert evaluation}
We invite a finance expert to inspect generated domain-specific nodes and reasoning cases on DeepFund. The expert found that Unified-MAS produces a reasonable investment-committee-style workflow, with multiple specialist agents, adversarial review, conflict resolution, and final decision synthesis. The expert also noted three strengths: the traces are structured and auditable, the insider-transaction analysis captures important transaction types, and the technical-analysis node produces actionable support/resistance and volume-based signals. However, the multi-agent system may produce misinformation, especially when deployed in real-world high-stakes domains such as healthcare, law, finance, or education. In practical deployment, we recommend that Unified-MAS should be intended to assist domain experts rather than replace human oversight in high-risk decision-making.
}

\subsection{Discussion}
\label{appendix:discussion}

\emnlpcr{
At the component level, Unified-MAS involves retrieval, node/prompt synthesis, and validation-based optimization. However, our contribution is to introduce a new mechanism for Automatic-MAS: decoupling domain-specific node implementation from topology orchestration. This is different from standard RAG, prompt optimization, or tool-augmented agents, where the central goal is usually to improve a single agent or a fixed workflow, rather than to relieve an Automatic-MAS orchestrator from the simultaneous design of domain-specific node logic and the search for the global MAS topology.
}

\arrmay{For scalability, Unified-MAS scales effectively to highly complex domains that require a large number of specialized agents. As task complexity increases, standard Automatic-MAS with pre-defined general nodes typically fail to construct meaningful workflows. Similarly, dynamic generation methods struggle because handling both expert node design and topology orchestration over long horizons overwhelms the LLM. By decoupling node generation, Unified-MAS reliably synthesizes a targeted team of domain experts offline, enabling the orchestrator to scale to complex, long-chain multi-agent tasks.}

\arrmay{To solve the real-time adaptation problem in the future, we could introduce a dynamic monitoring mechanism to the framework. For example, if the deployed system encounters an entirely new scenario during online inference (such as a newly enacted financial regulation or a novel medical virus), an ``online monitor'' could detect the execution failure or low confidence. It could then temporarily trigger a lightweight, real-time version of our search-based generation module to synthesize a new node on the fly, without waiting for a full offline update. Additionally, live execution feedback could be continuously cached to perform rapid, localized node refinement.}

\subsection{Further Ablation on Hyperparameters}
\label{appendix:further ablation}
\arrmay{
To investigate the effect of some key hyperparameters and setups, we further conduct four ablation studies in Table~\ref{tab:ablation-ds-gpt}, i.e., Retrieval source, maximum search turns (Turn), reward weighting factor ($\alpha$), and size of validation set.
}

\arrmay{
\textbf{Retrieval Sources}: Our three retrieval sources (Google, GitHub, Google Scholar) are complementary and are designed to search different types of external information. Here, we remove the Google Scholar source, and performance noticeably drops, confirming the necessity of combining diverse sources.
}

\arrmay{
\textbf{Hyperparameters}: we test the sensitivity to the maximum search turns (Turn$= 5, 15$), the reward weighting factor ($\alpha = 0.2, 0.4, 0.8$), and validation set size (using only half the set). 
Our original setting is Full-validation set, Turn$= 10$, and $\alpha = 0.6$. The performance fluctuations are generally minimal across different maximum search turns, reward weighting factors, and validation set sizes. This demonstrates the overall robustness of our framework. For the weighting factor $\alpha$, the performance remains largely stable around our default setting. We only observe a performance drop in a specific extreme case (i.e., AFlow evaluated with DeepSeek-V3.2 when $\alpha=0.2$). This occasional variance is expected, as extreme values can disrupt the intended balance between the improvement score and the consistency score. Overall, these results confirm that Unified-MAS is robust and not overly sensitive to hyperparameter choices in most scenarios.
}

\begin{table}[htbp]
    \centering
    \resizebox{0.48\textwidth}{!}{%
    \begin{tabular}{@{}l|cc|cc@{}}
    \toprule
     \multirow{2.5}{*}{Variants} & \multicolumn{2}{c|}{DeepSeek-V3.2} & \multicolumn{2}{c}{GPT-5-Mini} \\
     \cmidrule(lr){2-3} \cmidrule(l){4-5}
    & AFlow & MAS$^2$ & AFlow & MAS$^2$ \\ \midrule
    Ours & 45.79 & 28.97 & 44.86 & 36.45 \\\midrule
    w/o Google Scholar & 40.19 & 23.36 & 39.25 & 30.84 \\\midrule
    Search Turn = 5 & 43.93 & 25.23 & 41.12 & 32.71 \\
    Search Turn = 15 & 44.86 & 28.04 & 42.06 & 33.64 \\\midrule
    $\alpha = 0.2$ & 39.25 & 26.17 & 42.06 & 32.71 \\
    $\alpha = 0.4$ & 42.06 & 28.04 & 45.79 & 34.58 \\
    $\alpha = 0.8$ & 42.99 & 24.30 & 43.93 & 31.78 \\\midrule
    Half-validation set & 46.73 & 27.10 & 42.99 & 35.51 \\
    \bottomrule
    \end{tabular}%
    }

    \caption{Ablation study results evaluated on DeepFund with Gemini-3-Flash as the Designer.}
    \label{tab:ablation-ds-gpt}
\end{table}

\emnlpcr{
\subsection{More Comparison Experiments}
As shown in Table~\ref{tab:more-comparison}, we add more ablation studies on the retrieval source and reward choices, i.e., random bottleneck-node selection, reward-model computation~\cite{dubey2024llama}, and reward stability under randomized node orders.
On one hand, the result shows that all three external sources contribute, while the full Unified-MAS setting performs best.
On the other hand, it further suggests that the perplexity-guided reward is effective and reasonably stable: randomized node orders preserve most of the gain, while random bottleneck selection and reward-model computation (Llama-3-8b-rm-mixture) underperform the default design.
}

\begin{table}[htbp]
    \centering
    \resizebox{0.48\textwidth}{!}{%
    \begin{tabular}{@{}l|c@{}}
    \toprule
     Variants & DeepSeek-V3.2 \\ \midrule
    MAS$^2$ & \textbf{28.97} \\ \midrule
    MAS$^2$ + no retrieval & 24.30 \\
    MAS$^2$ + Google only & 27.10 \\
    MAS$^2$ + Scholar only & 26.16 \\
    MAS$^2$ + GitHub only & 25.23 \\\midrule
    MAS$^2$ + random bottleneck selection & 23.36 \\
    MAS$^2$ + reward-model computation & 21.50 \\
    MAS$^2$ + reward stability under randomized orders & 27.10 \\
    \bottomrule
    \end{tabular}%
    }

    \caption{More comparison results evaluated on DeepFund with Gemini-3-Flash as the Designer.}
    \label{tab:more-comparison}
\end{table}

\emnlpcr{
In Table~\ref{tab:equip-baseline-search}, we equip MetaAgent, EvoAgent, and AOrchestra with RAG (online searching) as an additional sub-agent, and further evaluate self-consistency with three runs to approximate the offline budget. The MAS$^2$'s results on DeepFund with DeepSeek-V3.2 show that both RAG and self-consistency improve the baselines, but they still remain below Unified-MAS. This suggests that the offline domain-specific node library generated by Unified-MAS provides a stronger improvement than directly attaching RAG to dynamic-generation baselines.
}

\begin{table}[htbp]
    \centering
    \begin{tabular}{@{}l|ccc@{}}
    \toprule
     Methods & Original & + RAG & + SC=3\\ \midrule
    EvoAgent & 28.97 & 30.84 & 36.45 \\
    MetaAgent & 33.64 & 35.51 & 39.25 \\
    AOrchestra & 25.23 & 28.97 & 29.91\\
    Unified-MAS best & \textbf{46.73} & - & - \\
    \bottomrule
    \end{tabular}%

    \caption{Comparison against stronger dynamic Automatic-MAS baselines evaluated on MAS$^2$, DeepFund with Gemini-3-Flash as the Designer.}
    \label{tab:equip-baseline-search}
\end{table}

\section{Case Study}
\label{appendix:case study}

Table~\ref{tab:aime nodes} lists the generated nodes of Unified-MAS using Gemini-3-Pro on TravelPlanner, HealthBench, DeepFund, and AIME24\&25. For further comparison, Table~\ref{tab:generated nodes} shows the generated nodes using Unified-MAS and other Automatic-MAS with dynamic nodes. It indicates that although these Automatic-MAS can introduce new nodes to some extent, their performance in different specialized fields is not stable enough. For example, EvoAgent generates an excessive number of ``Expert Node'' to solve the problem in parallel, which is more like an ensemble rather than introducing the real agentic element.



\begin{table*}[htbp]
    \centering
    \footnotesize
    \resizebox{\textwidth}{!}{%
    \begin{tabular}{@{}lc@{\hspace{1.5em}}lc@{}}
    \toprule
    \multicolumn{2}{c}{\textbf{DeepFund}} & \multicolumn{2}{c}{\textbf{AIME24\&25}} \\
    \cmidrule(lr){1-2} \cmidrule(lr){3-4}
    Node Name & Node Description & Node Name & Node Description \\
    \midrule
    Fundamental\_Analysis\_Agent & \begin{tabular}[c]{@{}c@{}}Analyzes news and policy text\\ for sentiment and key drivers.\end{tabular} &
    Math\_Domain\_Analyzer & \begin{tabular}[c]{@{}c@{}}Understands the problem\\ type and key constraints.\end{tabular} \\
    \addlinespace[2pt]
    Insider\_Analysis\_Agent & \begin{tabular}[c]{@{}c@{}}Detects insider trading patterns\\ and conviction signals.\end{tabular} &
    Theorem\_Strategy\_Retriever & \begin{tabular}[c]{@{}c@{}}Finds relevant theorems\\ and solving strategies.\end{tabular} \\
    \addlinespace[2pt]
    Technical\_Analysis\_Agent & \begin{tabular}[c]{@{}c@{}}Interprets OHLCV data for trend,\\ support, and resistance levels.\end{tabular} &
    Step\_by\_Step\_Solver & \begin{tabular}[c]{@{}c@{}}Builds a full solution\\ draft step by step.\end{tabular} \\
    \addlinespace[2pt]
    Risk\_Management\_Agent & \begin{tabular}[c]{@{}c@{}}Checks portfolio cash, exposure,\\ and trading constraints.\end{tabular} &
    Constraint\_Logic\_Verifier & \begin{tabular}[c]{@{}c@{}}Checks and fixes\\ logic/math mistakes.\end{tabular} \\
    \addlinespace[2pt]
    CIO\_Synthesis\_Agent & \begin{tabular}[c]{@{}c@{}}Weighs all signals to produce\\ the final trading action.\end{tabular} &
    Final\_Answer\_Formatter & \begin{tabular}[c]{@{}c@{}}Extracts and formats\\ the final answer correctly.\end{tabular} \\
    \midrule
    \multicolumn{2}{c}{\textbf{HealthBench}} & \multicolumn{2}{c}{\textbf{TravelPlanner}} \\
    \cmidrule(lr){1-2} \cmidrule(lr){3-4}
    Node Name & Node Description & Node Name & Node Description \\
    \midrule
    Interaction\_Analyzer & \begin{tabular}[c]{@{}c@{}}Parses conversation history for\\ intent and medical entities.\end{tabular} &
    Requirement\_Parser & \begin{tabular}[c]{@{}c@{}}Extracts dates, budget, party size,\\ and locations from query.\end{tabular} \\
    \addlinespace[2pt]
    Guideline\_Retriever & \begin{tabular}[c]{@{}c@{}}Retrieves clinical guidelines\\ and red flag protocols via RAG.\end{tabular} &
    Logistics\_Planner & \begin{tabular}[c]{@{}c@{}}Selects flights and hotels\\ within budget constraints.\end{tabular} \\
    \addlinespace[2pt]
    Clinical\_Reasoner & \begin{tabular}[c]{@{}c@{}}Applies ESI triage logic and\\ generates draft responses.\end{tabular} &
    Daily\_Activity\_Scheduler & \begin{tabular}[c]{@{}c@{}}Schedules daily meals and\\ attractions in remaining budget.\end{tabular} \\
    \addlinespace[2pt]
    Safety\_Validator & \begin{tabular}[c]{@{}c@{}}Verifies safety compliance and\\ hedges diagnostic language.\end{tabular} &
    Final\_Formatter & \begin{tabular}[c]{@{}c@{}}Compiles the plan into the\\ required JSON itinerary format.\end{tabular} \\
    \bottomrule
    \end{tabular}%
    }
    \caption{\arrmay{Unified-MAS's generated nodes using Gemini-3-Pro across four benchmarks.}}
    \label{tab:aime nodes}
\end{table*}

Figure~\ref{fig:the mas with unified-MAS} and Figure~\ref{fig:the mas without unified-MAS} compare the different MAS generated by MAS-Zero using Gemini-3-Flash, for the same example shown in Table~\ref{tab:generated nodes}, with/without nodes generated by Unified MAS. Compared with the original AFlow, the Unified-MAS version is more structured, transparent, and reliable. It explicitly separates case structuring, legal retrieval, fact verification, damages calculation, and judgment drafting, so each reasoning stage is traceable and easier to validate. By contrast, the original AFlow relies more on prompt-level reasoning and ensemble voting, offering less explicit alignment between evidence, legal rules, and quantified outcomes.

\begin{table*}[htbp]
    \centering

    \setlength{\tabcolsep}{4.5pt}
    \footnotesize
    
    \resizebox{\textwidth}{!}{%
    \begin{tabular}{>{\RaggedRight\arraybackslash}p{0.13\textwidth}
                    >{\RaggedRight\arraybackslash}p{0.35\textwidth}
                    >{\RaggedRight\arraybackslash}p{0.48\textwidth}}
        \toprule
        \textbf{Method} & \textbf{Node Name} & \textbf{Node Description / Function} \\
        \midrule
        \rowcolor{gray!12}
        \multicolumn{3}{p{\textwidth}}{
            \textbf{Input Question:}
            You are a rigorous and impartial presiding judge. Your task is to generate legal reasoning and deliver the final ruling based on the plaintiff's and defendant's statements and the additional information provided. Maintain a neutral, professional, and fair judicial tone at all times, without favoring either side. You are given the following information:
            \texttt{\{"category": "Personality rights dispute", "plaintiff": "xx Song", "defendant": "A kindergarten in Beijing",}
            \texttt{"incident": "2023-04-21 kite activity injury (facial cut near eye)", "claims":[medical 6900¥, lost wages 4000¥, transport 3000¥,}
            \texttt{mental distress 10000¥, future treatment 40000¥], ...\}}
        } \\
        \midrule

        \multirow{6}{*}{\begin{tabular}[c]{@{}c@{}}\textbf{AOrchestra}\\ (Gemini-3-Flash)\end{tabular}}
        & \texttt{MainAgent} & Top-level orchestrator deciding next action. \\
        & \texttt{error} & Runtime error transition captured in trajectory log. \\
        & \texttt{delegate\_task} & Delegates current sub-problem to a sub-agent. \\
        & \texttt{finish} & Sub-agent final answer step for delegated task. \\
        & \texttt{complete} & MainAgent composes and returns final answer. \\
        & \texttt{SubAgent} & Delegated worker agent that executes subtask reasoning. \\
        \midrule

        \multirow{7}{*}{\begin{tabular}[c]{@{}c@{}}\textbf{EvoAgent}\\ (Gemini-3-Flash)\end{tabular}}
        & \texttt{MainAgent} & Controls iterative expert evolution and selection. \\
        & \texttt{Expert\#1} & Expert role (tort-law doctrine and social public policy). \\
        & \texttt{Expert\#2} & Expert role (protection of minors' rights and mental-health assessment). \\
        & \texttt{Expert\#3} & Expert role for refining disputed issues (future treatment and long-term impact). \\
        & \texttt{ExpertGroup(3)} & Aggregated 3-expert panel output per iteration. \\
        \midrule

        \multirow{3}{*}{\begin{tabular}[c]{@{}c@{}}\textbf{MetaAgent}\\ (Gemini-3-Flash)\end{tabular}}
        & \texttt{Presiding\_Judge} & Performs legal analysis: statute search, liability split, claim acceptance/rejection, and summary for downstream actuarial calculation. \\
        \midrule

        \multirow{12}{*}{\begin{tabular}[c]{@{}c@{}}\textbf{Unified\_MAS}\\ (Gemini-3-Flash)\end{tabular}}
        & \texttt{Rhetorical\_Segmenter} & Segments legal input into modules: plaintiff claims, defense, findings, and evidence. \\
        & \texttt{Legal\_Element\_Extractor} & Extracts legal-technical elements such as claim items, amounts, and injury/contract details. \\
        & \texttt{Statutory\_Retriever} & Retrieves applicable PRC Civil Code statutes based on extracted legal elements. \\
        & \texttt{Evidence\_Evaluator} & Evaluates evidentiary support using civil ``high probability'' proof standard. \\
        & \texttt{Liability\_Reasoning\_Engine} & Applies law to verified facts to infer liability ratio and compensation basis. \\
        & \texttt{Final\_Judgement\_Synthesizer} & Produces final judicial reasoning and verdict in required output format. \\
        \midrule

        \multirow{21}{*}{\begin{tabular}[c]{@{}c@{}}\textbf{Unified\_MAS}\\ (GPT-5-Mini)\end{tabular}}
        & \texttt{Ingest\_and\_Normalize} & Normalizes input into canonical text blocks with metadata and offsets. \\
        & \texttt{Document\_Classifier} & Classifies document/domain type and extracts top remedies. \\
        & \texttt{Party\_and\_Role\_Extraction} & Extracts parties/roles with provenance. \\
        & \texttt{Claims\_and\_Remedies\_Extraction} & Extracts requested claims/remedies and maps them to claimants. \\
        & \texttt{Evidence\_Enumeration} & Enumerates/classifies evidence and links evidence to claims/events. \\
        & \texttt{Timeline\_and\_Causation} & Builds a chronological event timeline and causal links to damages. \\
        & \texttt{Retrieve\_Statutes\_and\_Precedents} & Retrieves legal statutes and precedent snippets (RAG). \\
        & \texttt{Statute\_to\_Fact\_Linking} & Links facts/claims to statute or case references with justifications. \\
        & \texttt{Liability\_Reasoning} & Infers party liability allocation with legal rationale. \\
        & \texttt{Damage\_Calculation\_and\_Reconciliation} & Performs component-level damage calculation and reconciliation. \\
        & \texttt{Validation\_and\_Consistency\_Checks} & Runs consistency/constraint checks on full structured output. \\
        & \texttt{Final\_Judgment\_Synthesis} & Synthesizes full Chinese judgment text and structured verdict. \\
        & \texttt{Final\_Answer\_Line} & Emits the final one-line verdict beginning with ``Answer:''. \\
        \midrule

        \multirow{9}{*}{\begin{tabular}[c]{@{}c@{}}\textbf{Unified\_MAS}\\ (Gemini-3-Pro)\end{tabular}}
        & \texttt{Case\_Structurer} & Parses raw case JSON into parties, cause of action, claims, and dispute summary. \\
        & \texttt{Legal\_Search\_Engine} & Retrieves statutes/judicial interpretations relevant to the dispute type. \\
        & \texttt{Fact\_Analyzer} & Verifies facts and causality from conflicting statements and evidence. \\
        & \texttt{Damages\_Calculator} & Validates and computes monetary compensation items. \\
        & \texttt{Judgment\_Drafter} & Drafts the final formal judgment text from structured reasoning. \\

        \bottomrule
    \end{tabular}
    }

    \caption{Comparison of generated nodes using MetaAgent, EvoAgent, AOrchestra, and Unified-MAS on J1Bench.}
    \label{tab:generated nodes}
\end{table*}

\begin{table*}[htbp]
\centering
\setlength{\tabcolsep}{4.5pt}
\footnotesize
\resizebox{\textwidth}{!}{%
\begin{tabular}{>{\RaggedRight\arraybackslash}p{0.15\textwidth}
                >{\RaggedRight\arraybackslash}p{0.85\textwidth}}
\toprule
\textbf{Epoch} & \textbf{Node Internal Implementation} \\
\midrule

\multirow{9}{*}{\begin{tabular}[l]{@{}l@{}}\textbf{Epoch 0}\\ \textbf{(Initialization)}\end{tabular}}
&
\begin{minipage}[t]{\linewidth}
\texttt{def Fact\_Analyzer(self, input\_data):}\\
\texttt{\ \ \ \ args = input\_data.get("arguments", \{\})}\\
\texttt{\ \ \ \ evid = input\_data.get("evidence\_summary", \{\})}\\
\texttt{\ \ \ \ node\_messages = [}\\
\texttt{\ \ \ \ \ \ \ \ \{"role":"system","content":"You are a Senior Investigator..."\},}\\
\texttt{\ \ \ \ \ \ \ \ \{"role":"user","content":f"Plaintiff's Story \& Evidence: \{args\}, \{evid\}..."\}}\\
\texttt{\ \ \ \ ]}\\
\texttt{\ \ \ \ response = \colorbox{yellow!35}{\texttt{self.llm\_client.chat}}(node\_messages, response\_format="json\_object")}\\
\texttt{\ \ \ \ return response}
\end{minipage}
\\

\midrule

\multirow{12}{*}{\begin{tabular}[l]{@{}l@{}}\textbf{Epoch 10}\\ \textbf{(Optimized)}\end{tabular}}
&
\begin{minipage}[t]{\linewidth}
\texttt{def Fact\_Analyzer(self, input\_data):}\\
\texttt{\ \ \ \ \# omitted detailed prompt bodies for space} \\
\texttt{\ \ \ \ cause\_of\_action = input\_data.get('cause\_of\_action', 'Civil Dispute')}\\
\texttt{\ \ \ \ parties = input\_data.get('parties') or input\_data.get('specific\_characters', \{\})}\\
\texttt{\ \ \ \ arguments = input\_data.get('arguments', \{\})}\\
\texttt{\ \ \ \ evidence = input\_data.get('evidence', \{\})}\\
\texttt{\ \ \ \ evidence\_str = json.dumps(evidence, ensure\_ascii=False, indent=2)}\\
\texttt{\ \ \ \ fact\_finder\_messages = [system: category-aware rules for Tort/Financial/Divorce, user: case context]}\\
\texttt{\ \ \ \ fact\_analysis = \colorbox{yellow!35}{\texttt{self.llm\_client.chat}}(fact\_finder\_messages)}\\
\texttt{\ \ \ \ judge\_messages = [system: liability matrix, user: Fact Analysis]}\\
\texttt{\ \ \ \ response = \colorbox{yellow!35}{\texttt{self.llm\_client.chat}}(judge\_messages, response\_format='json\_object')}\\
\texttt{\ \ \ \ return response}\\

\end{minipage}
\\

\bottomrule
\end{tabular}%
}
\caption{Comparison of \texttt{Fact\_Analyzer} implementation across epochs on J1Bench. Compared to the unoptimized Epoch 0, Epoch 10 contains a two-stage, category-aware reasoning pipeline.}
\label{tab:strong node}
\end{table*}

\begin{figure*}[htbp]
    \centering
    \begin{tcolorbox}[
        enhanced,
        colframe=Salmon!90!Black,          
        colback=Salmon!20,         
        coltitle=white,         
        fonttitle=\large\bfseries, 
        title={The MAS generated by AFlow with Unified-MAS}, 
        halign title=left,      
        fontupper=\footnotesize,       
        boxrule=1pt,            
        arc=3mm,                
        boxsep=2pt,
        left=8pt,               
        right=8pt,              
        top=4pt,
        bottom=4pt
    ]

\hspace{0em}class Workflow:

\hspace{2em}def \_\_init\_\_(

\hspace{4em}self,

\hspace{4em}name: str,

\hspace{4em}llm\_config,

\hspace{4em}dataset: DatasetType,

\hspace{2em}) -> None:

\hspace{4em}self.name = name

\hspace{4em}self.dataset = dataset

\hspace{4em}self.llm = create\_llm\_instance(llm\_config)

\hspace{4em}self.custom = operator.Custom(self.llm)

\hspace{4em}self.case\_structurer = operator.CaseStructurer(self.llm)

\hspace{4em}self.fact\_analyzer = operator.FactAnalyzer(self.llm)

\hspace{4em}self.legal\_search\_engine = operator.LegalSearchEngine(self.llm)

\hspace{4em}self.damages\_calculator = operator.DamagesCalculator(self.llm)

\hspace{4em}self.judgment\_drafter = operator.JudgmentDrafter(self.llm)

\hspace{2em}async def \_\_call\_\_(self, problem: str):

\hspace{4em}"""

\hspace{4em}Implementation of a structured judicial workflow using specialized operators.

\hspace{4em}"""

\hspace{4em}\textbf{\# Step 1: Extract and structure the key legal elements from raw case text}

\hspace{4em}s\_res = await self.case\_structurer(input\_data=problem)

\hspace{4em}\# Prepare inputs for fact analysis (ensuring dictionary types for keys expected to be objects)

\hspace{4em}fact\_arguments = s\_res.get("arguments", \{\})

\hspace{4em}fact\_parties = s\_res.get("parties", \{\})        

\vspace{5pt}

\hspace{4em}\textbf{\# Step 2: Retrieve relevant laws and perform forensic fact analysis}

\hspace{4em}l\_res = await self.legal\_search\_engine(cause\_of\_action=str(s\_res.get("cause\_of\_action", "")), arguments=str(s\_res.get("arguments", "")))

\hspace{4em}\# Ensure legal search result is a dictionary before accessing "relevant\_laws"

\hspace{4em}relevant\_laws = l\_res.get("relevant\_laws", str(l\_res)) if isinstance(l\_res, dict) else str(l\_res)

\hspace{4em}f\_res = await self.fact\_analyzer(

\hspace{6em}arguments=fact\_arguments, 

\hspace{6em}evidence\_summary=str(s\_res.get("evidence\_summary", "")), 

\hspace{6em}cause\_of\_action=str(s\_res.get("cause\_of\_action", "")), 

\hspace{6em}parties=fact\_parties

\hspace{4em})
        
\vspace{5pt}
\hspace{4em}\textbf{\# Step 3: Calculate valid damages based on verified facts and legal rules}

\hspace{4em}c\_res = await self.damages\_calculator(

\hspace{6em}financial\_claims=s\_res.get("financial\_claims", []), 

\hspace{6em}verified\_facts=f\_res, 

\hspace{6em}relevant\_laws=relevant\_laws, 

\hspace{6em}cause\_of\_action=str(s\_res.get("cause\_of\_action", ""))

\hspace{4em})
        
\vspace{5pt}
\hspace{4em}\textbf{\# Step 4: Generate a formal draft judgment}

\hspace{4em}dr\_res = await self.judgment\_drafter(

\hspace{6em}structured\_brief=s\_res, 

\hspace{6em}relevant\_laws=relevant\_laws, 

\hspace{6em}verified\_facts=f\_res, 

\hspace{6em}damages\_table=c\_res

\hspace{4em})
        
\vspace{5pt}
\hspace{4em}\textbf{\# Step 5: Final synthesis and rule enforcement (Statute of Limitations, Divorce grounds, Causality)}

\hspace{4em}\# Safely extract the judgment text from the drafting operator result

\hspace{4em}draft\_judgment = dr\_res.get("final\_judgment", str(dr\_res)) if isinstance(dr\_res, dict) else str(dr\_res)

\hspace{4em}context = f"Original Case: \{problem\},

\hspace{8em}Verified Facts: \{f\_res\},

\hspace{8em}Legal Draft: \{draft\_judgment\}"

\hspace{4em}solution = await self.custom(input=context, instruction=prompt\_custom.JUDGE\_PROMPT)

\hspace{4em}\# Handle final solution format

\hspace{4em}if isinstance(solution, dict):

\hspace{6em}final\_response = solution.get("response", str(solution))

\hspace{4em}else:

\hspace{6em}final\_response = str(solution)

\hspace{4em}return final\_response, self.llm.get\_usage\_summary()["total\_cost"]

    \end{tcolorbox}
    
    \caption{The MAS generated by AFlow with Unified-MAS using Gemini-3-Flash as Orchestrator.}
    \label{fig:the mas with unified-MAS}
\end{figure*}

\begin{figure*}[htbp]
    \centering
    \begin{tcolorbox}[
        enhanced,
        colframe=Salmon!90!Black,          
        colback=Salmon!20,         
        coltitle=white,         
        fonttitle=\large\bfseries, 
        title={The MAS generated by AFlow without Unified-MAS}, 
        halign title=left,      
        fontupper=\footnotesize,       
        boxrule=1pt,            
        arc=3mm,                
        boxsep=2pt,
        left=8pt,               
        right=8pt,              
        top=4pt,
        bottom=4pt
    ]

\hspace{0em}class Workflow:

\hspace{2em}def \_\_init\_\_(

\hspace{4em}self,

\hspace{4em}name: str,

\hspace{4em}llm\_config,

\hspace{4em}dataset: DatasetType,

\hspace{2em}) -> None:

\hspace{4em}self.name = name

\hspace{4em}self.dataset = dataset

\hspace{4em}self.llm = create\_llm\_instance(llm\_config)

\hspace{4em}self.custom = operator.Custom(self.llm)

\hspace{4em}self.sc\_ensemble = operator.ScEnsemble(self.llm)

\hspace{4em}self.answer\_generate = operator.AnswerGenerate(self.llm)

\hspace{2em}async def \_\_call\_\_(self, problem: str):

\hspace{4em}"""

\hspace{4em}Implementation of the workflow with Evidence checking, Logical accounting, and Ensemble judgment.

\hspace{4em}"""

\hspace{4em}\textbf{\# Step 1: Objective evaluation of the evidentiary chain and proof status}

\hspace{4em}evidence = await self.custom(input=problem, instruction=prompt\_custom.EVIDENCE\_PROMPT)

\hspace{4em}evidence\_res = evidence.get("response", "")
        
\vspace{5pt}
\hspace{4em}\textbf{\# Step 2: Step-by-step logical accounting of liability and proven amounts}

\hspace{4em}\# We use AnswerGenerate to enforce a CoT structure (thought) and a summary (answer)

\hspace{4em}logic = await self.answer\_generate(input=f"Evidence review result: \{evidence\_res\}, Case information: \{problem\}, Please strictly deduct amounts not admitted according to the \"dispute\" field, then analyze the liability ratio based on the defendant's defense and calculate the final amount.")

\hspace{4em}\# Safe extraction to prevent KeyError: "answer" if the LLM output doesn't match the expected tags

\hspace{4em}logic\_thought = logic.get("thought", "No detailed logical analysis")

\hspace{4em}logic\_answer = logic.get("answer", logic.get("response", "No clear core conclusion"))
        
\vspace{5pt}
\hspace{4em}\textbf{\# Step 3: Ensemble multiple candidates to reach the most consistent final judgment}

\hspace{4em}candidates = []

\hspace{4em}for \_ in range(3):

\hspace{6em}\# We pass both the detailed reasoning and the specific conclusion to the judge

\hspace{6em}cand\_res = await self.custom(

\hspace{8em}input=f"Logical analysis: \{logic\_thought\},

\hspace{8em}Core conclusion: \{logic\_answer\},

\hspace{8em}Case information: \{problem\}", 

\hspace{8em}instruction=prompt\_custom.JUDGMENT\_PROMPT

\hspace{6em})

\hspace{6em}candidates.append(cand\_res.get("response", ""))

\hspace{4em}\# Perform self-consistency ensemble to select the most reliable verdict

\hspace{4em}solution = await self.sc\_ensemble(problem=problem, solutions=candidates)

\hspace{4em}return solution.get("response", ""), self.llm.get\_usage\_summary()["total\_cost"]

    \end{tcolorbox}
    
    \caption{The MAS generated by AFlow without Unified-MAS using Gemini-3-Flash as Orchestrator.}
    \label{fig:the mas without unified-MAS}
\end{figure*}

\section{Prompt Details}
\label{appendix:prompt details}

We elaborate on the prompts used in Unified-MAS from Figure~\ref{fig:prompt for llm-as-a-judge evaluation} to Figure~\ref{fig:prompt for node optimization}. These comprehensive instructions cover evaluation and the entire framework pipeline, including keyword extraction, search query generation, strategy analysis, node generation, and node optimization.

\begin{figure*}[htbp]
    \centering
    \begin{tcolorbox}[
        enhanced,
        colframe=Salmon!90!Black,          
        colback=Salmon!20,         
        coltitle=white,         
        fonttitle=\large\bfseries, 
        title={Prompt for LLM-as-a-judge Evaluation}, 
        halign title=left,      
        fontupper=\footnotesize,       
        boxrule=1pt,            
        arc=3mm,                
        boxsep=2pt,
        left=8pt,               
        right=8pt,              
        top=4pt,
        bottom=4pt
    ]
    
    \textbf{HealthBench:}

    Your job is to score a model's reply against a rubric item. Decide whether the reply meets the criterion described in the rubric item.

    Model's reply:
    <reply>
    
    Rubric item:
    <rubric\_item>
    
    \# Instructions
    
    Return a json object with the following fields: "explanation" and "criteria\_met".
    
    The "explanation" field should be a string explaining why the reply does or does not meet the criteria of the rubric item.
    
    The "criteria\_met" field should be a boolean indicating whether the reply meets the criteria of the rubric item. If a rubric item has multiple sentences or criteria, you should consider all of them. If any of the criteria is not met, the answer should be false. Only return true if all of the criteria are met.
    
    One important exception: if a criteria says "such as", "for example", or "including", the reply does not have to include all of the examples listed to meet the criteria.
    
    \# Final instruction
    
    Return just the json object in markdown format. Do not include any other text in the response.

    \vspace{10pt}
    \textbf{J1Bench:}

    You are a legal expert. Please determine whether the "Answer to be evaluated" fully and accurately covers all the core points of the "Standard answer" based on the "legal issue" and the "Standard Answer", and then give a score ranging from 0 to 10. You don't need to consider non-substantive factors such as whether the answer to be evaluated is expressed concisely, whether the key points are highlighted, whether small talk is used, or whether the structure is lengthy. You don't need to deduct points for being insufficiently concise, nor do you need to consider the difference in language forms between Chinese and English. You only need to consider whether the content meaning is consistent.

    \textbf{Standard answer:}
    <gt\_answer>
    
    \textbf{Answer to be evaluated:}
    <mode\_answer>
    
    Output your results in the following format (no line breaks, no parentheses):
    <Rating: ..., Reason: ...>

    \end{tcolorbox}
    
    \caption{Prompt for LLM-as-a-judge Evaluation.}
    \label{fig:prompt for llm-as-a-judge evaluation}
\end{figure*}

\begin{figure*}[htbp]
    \centering
    \begin{tcolorbox}[
        enhanced,
        colframe=Salmon!90!Black,          
        colback=Salmon!20,         
        coltitle=white,         
        fonttitle=\large\bfseries, 
        title={Prompt for Keyword Extraction}, 
        halign title=left,      
        fontupper=\footnotesize,       
        boxrule=1pt,            
        arc=3mm,                
        boxsep=2pt,
        left=8pt,               
        right=8pt,              
        top=4pt,
        bottom=4pt
    ]

    \textbf{System\_prompt:}
    
    You are an expert dataset and task analyst. You are given multiple samples from a benchmark dataset. Your task is to carefully read all samples, analyze this Specific Domain Task, and extract keywords across six specific dimensions required to solve this task. The extracted keywords should be concise and representative, and should not focus on the specific data samples, but on the general domain and task.

    \textbf{User\_prompt:}
    
    User Task samples:
    <samples\_text>
    
    Analyze the description above and reason to extract keywords for the following six dimensions. For each dimension, provide 5-10 most representative terms:

    1.  \textbf{Domain:} The macro industry background (e.g., Fintech, Supply Chain, Bioinformatics, etc.).
    
    2.  \textbf{Task:} The core technical problem to solve (e.g., Anomaly Detection, Named Entity Recognition, Summarization, etc.).
    
    3.  \textbf{Entities:} The specific data objects or physical entities involved (e.g., Transaction Logs, PDF Contracts, Protein Sequences, Sensor Data, etc.).
    
    4.  \textbf{Actions:} The specific operations performed on the data (e.g., Classify, Extract, Reason, Optimize, Verify, etc.).
    
    5.  \textbf{Constraints:} Performance metrics or limitations (e.g., Low Latency, Privacy Preserving, Explainability, Offline Inference, etc.).
    
    6.  \textbf{Desired Outcomes:} The expected results or metrics (e.g., Accuracy, Precision, Recall, F1 Score, AUC, MAP, NDCG, etc.).
    
    7.  \textbf{Implicit Knowledge:} Based on your expert knowledge, infer specific jargon, SOTA techniques, common challenges, or potential risks that are not explicitly mentioned but are essential for solving this problem (e.g., "Imbalanced Data" for fraud, "Hallucination" for GenAI, "Bullwhip Effect" for supply chain, etc.).

    \textbf{\# Output Format}
    
    Please output both your thinking and answer in the JSON format.
    
    "thinking" entry: [Your thinking process, how you arrive your answer]
    
    "answer" entry: [your answer in the JSON format]

    For the "thinking" entry, you need to first carefully read the <samples\_text>, summarize the task description, and then reason step by step to arrive at your answer.

    For the "answer" entry, please output a valid JSON object. Do not include any conversational filler or markdown formatting outside the JSON code block. Format as follows:

    \{\{
    
        \hspace{2em}"Domain": ["..."],
        
        \hspace{2em}"Task": ["..."],
        
        \hspace{2em}"Entities": ["..."],
        
        \hspace{2em}"Actions": ["..."],
        
        \hspace{2em}"Constraints": ["..."],
        
        \hspace{2em}"Desired\_Outcomes": ["..."],
        
        \hspace{2em}"Implicit\_Knowledge": ["..."]
        
    \}\}

    \end{tcolorbox}
    
    \caption{Prompt for Keyword Extraction.}
    \label{fig:prompt for keyword extraction}
\end{figure*}

\begin{figure*}[htbp]
    \centering
    \begin{tcolorbox}[
        enhanced,
        colframe=Salmon!90!Black,          
        colback=Salmon!20,         
        coltitle=white,         
        fonttitle=\large\bfseries, 
        title={Prompt for Search Query Generation}, 
        halign title=left,      
        fontupper=\footnotesize,       
        boxrule=1pt,            
        arc=3mm,                
        boxsep=2pt,
        left=8pt,               
        right=8pt,              
        top=4pt,
        bottom=4pt
    ]

    \textbf{System\_prompt:}
    
    You are an expert in Information Retrieval (IR) and Multi-Agent System Design. You know how to construct precise search queries to retrieve background knowledge, high-quality academic papers, code implementations, and industry Standard Operating Procedures (SOPs).
    
    \textbf{User\_prompt:}
    
    Based on the provided [Structured Keywords (Domain, Task, Entities, Actions, Constraints, Desired\_Outcomes, Implicit\_Knowledge)], apply four specific search strategies to generate a list of search queries for Google Scholar, GitHub, and General Web Search.

    \textbf{Structured Keywords JSON:}
    <keywords\_json\_str>

    Apply the following four strategies to construct your queries for each dimension:

    1.  \textbf{Strategy A: Background Knowledge}
    
        Logic: Domain + Implicit\_Knowledge 
        
        Aim: Use domain jargon to find background knowledge, cutting-edge solutions, theoretical frameworks, and surveys.

    2.  \textbf{Strategy B: High-quality Academic Papers about System Architecture (Workflow \& Design)}
    
        Logic: Task + Constraints 
        
        Aim: Find architectural designs (e.g., Router, Pipeline, Map-Reduce) that satisfy specific constraints (e.g., Privacy, Real-time).

    3.  \textbf{Strategy C: Technical Code Implementation}
    
        Logic: Entities + Actions 
        
        Aim: Find code repositories, libraries, or preprocessing tools for specific data types.

    4.  \textbf{Strategy D: Evaluation \& Metrics}
    
        Logic: Task + Desired\_Outcomes 
        
        Aim: Find standard datasets and quantitative metrics to evaluate the Agent's performance.

    \textbf{\# Output Instructions}
    
    Generate 5-10 search queries for EACH strategy. Use Boolean operators (AND, OR) where appropriate to optimize results.

    Please output ONLY a valid JSON object with the following structure:

    \{\{
    
      \hspace{2em}"strategy\_A": [
      
        \hspace{4em}\{\{"query": "...", "reasoning": "Using [Implicit Term] to find advanced patterns"\}\}
        
      \hspace{2em}],
      
      \hspace{2em}"strategy\_B": [
      
        \hspace{4em}\{\{"query": "...", "reasoning": "To find architectures satisfying [Constraint]"\}\}
        
      \hspace{2em}],
      
      \hspace{2em}"strategy\_C": [
      
        \hspace{4em}\{\{"query": "...", "reasoning": "To find tools for processing [Object] via [Action]"\}\}
        
      \hspace{2em}],
      
      \hspace{2em}"strategy\_D": [
      
        \hspace{4em}\{\{"query": "...", "reasoning": "To find benchmarks for [Outcome]"\}\}
        
      \hspace{2em}]
      
    \}\}

    \end{tcolorbox}
    
    \caption{Prompt for Search Query Generation.}
    \label{fig:prompt for search query generation}
\end{figure*}

\begin{figure*}[htbp]
    \centering
    \begin{tcolorbox}[
        enhanced,
        colframe=Salmon!90!Black,          
        colback=Salmon!20,         
        coltitle=white,         
        fonttitle=\large\bfseries, 
        title={Prompt for Multi-turn Search}, 
        halign title=left,      
        fontupper=\footnotesize,       
        boxrule=1pt,            
        arc=3mm,                
        boxsep=2pt,
        left=8pt,               
        right=8pt,              
        top=4pt,
        bottom=4pt
    ]

    \textbf{System\_prompt:}
    You are a web search controller. Your job is to decide, step by step, how to search the web so that the user can find content that matches the target description. At each round, you will see the target description and a summary of past searches and results, and you must decide whether more searching is needed. You MUST respond with a valid JSON object only, no extra text.

    \textbf{User\_prompt:}
    
    Target description: <target\_description>
    
    Search round: <round\_idx> / <max\_rounds>
    
    Past search rounds: <history\_str>
    
    Search engine context:
    
    Backend type: <engine\_type>
    
    Instructions:
        
    \textbf{Github\_engine\_hint:}
    
    Current search backend: GitHub repository search. You MUST construct queries that look like GitHub repo searches, NOT natural language questions. Focus on a few core keywords: domain, task, entities, and techniques. Prefer short keyword-style queries, optionally with GitHub qualifiers such as 'language:python', 'in:name,description,readme', 'stars:>10'. Avoid 'survey of', 'methods for', 'towards', or very long sentences in the query.

    \textbf{Scholar\_engine\_hint:}
    
    Current search backend: Google Scholar (academic papers). You should construct queries that look like paper titles or combinations of technical terms. It is good to include phrases like 'survey', 'review', 'state of the art' when searching for overviews. Focus on scientific keywords (task, domain, methodology) rather than implementation details.

    \textbf{Google\_engine\_hint:}
    
    Current search backend: general Google web search. You may mix natural language with key technical terms. Focus on retrieving background knowledge, blog posts, documentation, or tutorials relevant to the target description.
    
    \textbf{Your task in THIS round:}
    
    1. Carefully read the target description and past search results.
    
    2. Decide whether we already have enough information that clearly matches the target description.
    
    3. If yes, set "done": true and summarize the useful information we already have.
    
    4. If no, set "done": false and propose the NEXT web search query to run.
    
    \textbf{Output JSON schema (you must strictly follow):}
    
    \{\{
    
      \hspace{2em}"done": bool,                 // true if we already have enough matching information
      
      \hspace{2em}"need\_search": bool,          // whether to run another web search in this round
      
      \hspace{2em}"next\_query": str,            // the next search query to run (empty if done=true)
      
      \hspace{2em}"reasoning": str,             // your reasoning for this decision
      
      \hspace{2em}"summary": str                // if done=true, summarize what has been found and why it matches
      
    \}\}

    \end{tcolorbox}
    
    \caption{Prompt for Multi-turn Search.}
    \label{fig:prompt for multi-turn search}
\end{figure*}

\begin{figure*}[htbp]
    \centering
    \begin{tcolorbox}[
        enhanced,
        colframe=Salmon!90!Black,          
        colback=Salmon!20,         
        coltitle=white,         
        fonttitle=\large\bfseries, 
        title={Prompt for Strategy\_A Analysis}, 
        halign title=left,      
        fontupper=\footnotesize,       
        boxrule=1pt,            
        arc=3mm,                
        boxsep=2pt,
        left=8pt,               
        right=8pt,              
        top=4pt,
        bottom=4pt
    ]

    \textbf{System\_prompt:}
    
    You are an expert technical analyst. Your task is to analyze multiple documents (PDFs and TXTs) that were retrieved through a web search for background knowledge related to a specific task, and provide a comprehensive summary.
 
    \textbf{User\_prompt:} 
    
    Task Description (from task\_keywords thinking) <task\_thinking>

    Strategy: <strategy\_name>

    Documents Retrieved: <files\_text>

    \textbf{IMPORTANT:} 
    
    Please provide an EXTREMELY DETAILED and COMPREHENSIVE analysis. The more detailed, the better. Include specific examples, step-by-step explanations, concrete details, and thorough descriptions.

    Your task:
    
    1. Analyze all the documents above and identify which aspects/aspects they discuss the background knowledge related to the task described above. Be very specific and detailed about each aspect.

    2. Summarize the key background information that is needed to solve this task. Provide EXTREMELY DETAILED descriptions, including but not limited to:
    
   \hspace{2em}\textbf{Overall task workflow and processes:} Provide a DETAILED, step-by-step workflow with specific stages, decision points, inputs/outputs at each stage, and the complete process flow. Include concrete examples and detailed explanations of each step.
   
   \hspace{2em}\textbf{Key points and important considerations:} List ALL important points with detailed explanations, why they matter, and how they impact the task. Be thorough and comprehensive.
   
   \hspace{2em}\textbf{Domain-specific knowledge and terminology:} Provide detailed definitions, explanations, and context for each term. Include how these concepts relate to each other and their significance in the domain.
   
   \hspace{2em}\textbf{Relevant frameworks, methodologies, or approaches:} Describe each framework/methodology in DETAIL, including their components, how they work, when to use them, and their advantages/disadvantages. Provide specific examples.
   
   \hspace{2em}\textbf{Common challenges and solutions:} Detail each challenge with specific scenarios, root causes, and provide detailed solutions with step-by-step approaches. Include real-world examples.
   
   \hspace{2em}\textbf{Best practices and standards:} Provide detailed best practices with specific guidelines, checklists, and detailed explanations of why each practice is important.

    3. Provide a structured summary that clearly explains:
    
   \hspace{2em}What background knowledge aspects are covered in these documents (with detailed descriptions)
   
   \hspace{2em}What specific background information is needed to solve the task (be very specific and detailed)
   
   \hspace{2em}How this background knowledge relates to the task at hand (provide detailed connections and relationships)

Remember: 

The more detailed and comprehensive your analysis, the better. Include specific examples, detailed explanations, step-by-step processes, and thorough descriptions throughout.

Please provide a comprehensive and well-structured analysis in JSON format:

\{\{

    \hspace{2em}"aspects\_covered": ["detailed aspect1 with explanation", "detailed aspect2 with explanation", ...],
    
    \hspace{2em}"background\_information": \{\{
    
        \hspace{4em}"task\_workflow": "DETAILED step-by-step workflow with all stages, inputs/outputs, decision points, and complete process flow. Be extremely thorough.",
        
        \hspace{4em}"key\_points": ["detailed point1 with full explanation", "detailed point2 with full explanation", ...],
        
        \hspace{4em}"domain\_knowledge": "DETAILED explanation of domain-specific knowledge, terminology, concepts, and their relationships. Be comprehensive and thorough.",
        
        \hspace{4em}"frameworks\_methodologies": ["detailed framework1 with components and usage", "detailed framework2 with components and usage", ...],
        
        \hspace{4em}"challenges\_solutions": "DETAILED description of common challenges with specific scenarios, root causes, and detailed step-by-step solutions with examples.",
        
        \hspace{4em}"best\_practices": "DETAILED best practices with specific guidelines, checklists, and explanations of importance. Be comprehensive."
        
    \hspace{2em}\}\},
    
    \hspace{2em}"summary": "EXTREMELY DETAILED and comprehensive summary of the background knowledge, including all key points, detailed workflows, and thorough explanations..."
    
\}\}

    \end{tcolorbox}
    
    \caption{Prompt for Strategy\_A Analysis.}
    \label{fig:prompt for strategy_a analysis}
\end{figure*}

\begin{figure*}[htbp]
    \centering
    \begin{tcolorbox}[
        enhanced,
        colframe=Salmon!90!Black,          
        colback=Salmon!20,         
        coltitle=white,         
        fonttitle=\large\bfseries, 
        title={Prompt for Strategy\_B Analysis}, 
        halign title=left,      
        fontupper=\footnotesize,       
        boxrule=1pt,            
        arc=3mm,                
        boxsep=2pt,
        left=8pt,               
        right=8pt,              
        top=4pt,
        bottom=4pt
    ]

    \textbf{System\_prompt:}
    
    You are an expert system architect and technical analyst. Your task is to analyze academic papers and documents about system architecture, workflow, and design related to a specific task, and provide insights on architectural patterns and design approaches.
        
    \textbf{User\_prompt:} 
    
    Task Description (from task\_keywords thinking): <task\_thinking>
    
    Strategy: <strategy\_name>
    
    Documents Retrieved: <files\_text>
    
    \textbf{IMPORTANT:} 
    
    Please provide an EXTREMELY DETAILED and COMPREHENSIVE analysis. The more detailed, the better. Include specific architectural diagrams, descriptions, detailed workflow steps, component interactions, and thorough explanations.
    
    Your task:
    
    1. Analyze all the documents above and identify the system architectures, workflows, and design patterns they discuss. Be very specific and detailed about each pattern and architecture.
    
    2. Summarize the key architectural and design information relevant to solving this task. Provide EXTREMELY DETAILED descriptions, including but not limited to:
    
       \hspace{2em}\textbf{System architecture patterns and structures:} Provide DETAILED descriptions of each architecture pattern, including components, their roles, data flow, communication patterns, and how they work together. Include specific examples and detailed explanations.
       
       \hspace{2em}\textbf{Workflow designs and process flows:} Provide EXTREMELY DETAILED, step-by-step workflow descriptions with all stages, transitions, decision points, data flows, error handling, and complete process flows. Include detailed diagrams, descriptions, and specific examples.
       
       \hspace{2em}\textbf{Component interactions and interfaces:} Detail how components interact, what interfaces they use, data formats, protocols, and communication mechanisms. Be very specific and thorough.
       
       \hspace{2em}\textbf{Design principles and constraints:} Provide detailed explanations of each design principle (e.g., privacy, real-time, scalability) with specific implementation strategies, trade-offs, and detailed guidelines. Include concrete examples.
       
       \hspace{2em}\textbf{Architectural trade-offs and decisions:} Detail each trade-off with specific scenarios, pros/cons, decision criteria, and detailed explanations of why certain choices are made. Be comprehensive.
       
       \hspace{2em}\textbf{Best practices for system design:} Provide detailed best practices with specific guidelines, patterns to follow, anti-patterns to avoid, and detailed explanations. Include real-world examples.
    
    3. Provide a structured summary that clearly explains:
       
       \hspace{2em}What architectural patterns and workflows are covered in these documents (with detailed descriptions)
       
       \hspace{2em}What specific architectural/design information is needed to solve the task (be very specific and detailed)
       
       \hspace{2em}How these architectural approaches relate to the task requirements (provide detailed connections and relationships)
    
    Remember: 
    
    The more detailed and comprehensive your analysis, the better. Include specific architectural details, detailed workflow steps, component interactions, and thorough explanations throughout.
    
    Please provide a comprehensive and well-structured analysis in JSON format:
    
    \{\{
    
        \hspace{2em}"architectural\_patterns": ["detailed pattern1 with components and structure", "detailed pattern2 with components and structure", ...],
        
        \hspace{2em}"design\_information": \{\{
        
            \hspace{4em}"system\_architectures": "DETAILED description of system architectures with components, data flows, communication patterns, and how they work together. Be extremely thorough.",
            
            \hspace{4em}"workflow\_designs": ["DETAILED step-by-step workflow1 with all stages and transitions", "DETAILED step-by-step workflow2 with all stages and transitions", ...],
            
            \hspace{4em}"component\_interactions": "DETAILED description of component interactions, interfaces, data formats, protocols, and communication mechanisms. Be comprehensive.",
            
            \hspace{4em}"design\_constraints": ["detailed constraint1 with implementation strategies", "detailed constraint2 with implementation strategies", ...],
            
            \hspace{4em}"architectural\_tradeoffs": "DETAILED description of trade-offs with specific scenarios, pros/cons, decision criteria, and explanations. Be thorough.",
            
            \hspace{4em}"design\_best\_practices": "DETAILED best practices with specific guidelines, patterns, anti-patterns, and explanations. Include examples. Be comprehensive."
        
        \hspace{2em}\}\},
        
        \hspace{2em}"summary": "EXTREMELY DETAILED and comprehensive summary of the architectural and design knowledge, including all patterns, detailed workflows, and thorough explanations..."
        
    \}\}

    \end{tcolorbox}
    
    \caption{Prompt for Strategy\_B Analysis.}
    \label{fig:prompt for strategy_b analysis}
\end{figure*}

\begin{figure*}[htbp]
    \centering
    \begin{tcolorbox}[
        enhanced,
        colframe=Salmon!90!Black,          
        colback=Salmon!20,         
        coltitle=white,         
        fonttitle=\large\bfseries, 
        title={Prompt for Strategy\_C Analysis}, 
        halign title=left,      
        fontupper=\footnotesize,       
        boxrule=1pt,            
        arc=3mm,                
        boxsep=2pt,
        left=8pt,               
        right=8pt,              
        top=4pt,
        bottom=4pt
    ]

    \textbf{System\_prompt:}
    
    You are an expert AI system architect and LLM prompt engineer. Your task is to analyze code repositories and design frameworks for solving tasks using Large Language Models (LLMs). Focus on high-level architecture, operation design, and how to migrate traditional ML/small model approaches to LLM-based solutions.
        
    \textbf{User\_prompt:}
    
    Task Description (from task\_keywords thinking): <task\_thinking>
    
    Strategy: <strategy\_name>
    
    Documents Retrieved (Code Repositories): <files\_text>
    
    \textbf{IMPORTANT:} 
    Focus on FRAMEWORK DESIGN and LLM MIGRATION, not on specific libraries or dependencies. Think about how to solve the task at a high level using LLMs.
    
    Your task:
    
    1. Analyze the overall framework and architecture in the provided code: 
    
    What is the high-level workflow and operation flow?
       How are different components organized and connected?
       What are the key operations/steps needed to solve the task?
       How can these operations be efficiently designed and orchestrated?
    
    2. Design LLM-based solutions to replace or enhance the small model implementations: 
    
    \textbf{Operation Design:} How to break down the task into well-defined operations that can be executed by LLMs? What operations are needed and how should they be structured?
       \textbf{Prompt Engineering:} For each operation that was previously done by small models, design detailed prompts for LLMs. What should be the input format, what instructions should be given, and what output format is expected?
       \textbf{Model-level Mechanisms:} How to implement global constraint checking, validation, error handling, and other model-level controls? What mechanisms are needed to ensure the LLM operations work correctly together?
       \textbf{Data Flow:} What is the input/output format for each LLM operation? How should data flow between different operations? What transformations are needed?
    
    3. Migration Strategy: 
    
    How can the existing small model code be adapted to use LLMs instead?
       What are the key differences in approach between small models and LLMs for this task?
       How to design the system to leverage LLM capabilities while maintaining the original workflow structure?
    
    4. Framework Considerations: 
    
    What is the overall system architecture needed to solve this task?
       How should operations be orchestrated and sequenced?
       What are the critical decision points and branching logic?
       How to handle state management and context passing between operations?
    
    \textbf{Focus Areas} (in order of importance):
    
    (1) Overall Framework \& Architecture: How to structure the solution at a high level.
    (2) Operation Design: How to break down the task into LLM-executable operations.
    (3) Prompt Design: Detailed prompt templates for each LLM operation.
    (4) Data Processing \& Flow: Input/output formats and data transformations between operations.
    (5) Model-level Mechanisms: Global constraints, validation, error handling.
    (6) Migration Strategy: How to adapt small model code to LLM-based approach.
    
    \textbf{Do NOT focus on:} Specific library dependencies or installation requirements.
    Environment setup details.
    Low-level implementation details of non-LLM components.
    
    Please provide a comprehensive and well-structured analysis in JSON format:
    
    \{\{
    
        \hspace{2em}"overall\_framework": \{\{
            "architecture": "DETAILED description of the overall system architecture and framework design needed to solve this task. Explain the high-level structure, component organization, and how different parts work together.",
            "workflow": "DETAILED step-by-step workflow description. Explain the sequence of operations, decision points, and how the system processes the task from start to finish.",
            key\_operations": ["operation1: detailed description of what it does and how it fits in the framework", "operation2: ...", ...]
        \}\},
        
        \hspace{2em}"llm\_migration": \{\{
            "operation\_design": "DETAILED description of how to design operations for LLM execution. Explain how to break down the task into operations, how operations should be structured, and how they should interact.",
            "prompt\_templates": [
                \{\{
                    "operation\_name": "name of the operation",
                    "purpose": "what this operation does in the overall framework",
                    "input\_format": "detailed description of input format and structure",
                    "prompt\_template": "detailed prompt template with placeholders and instructions",
                    "output\_format": "detailed description of expected output format",
                    "constraints": "any constraints or validation rules for this operation"
                \}\},
                ...
            ],
            "model\_level\_mechanisms": "DETAILED description of model-level mechanisms needed: global constraint checking, validation rules, error handling strategies, state management, context passing, etc. Be very specific about how these mechanisms work.",
            "migration\_strategy": "DETAILED explanation of how to migrate from small model code to LLM-based approach. What changes are needed, what can be reused, and how to adapt the existing workflow."
        \}\},
        
        \hspace{2em}"data\_processing": \{\{
            "input\_output\_formats": "DETAILED description of input/output formats for LLM operations. What data structures are needed, what format should be used, and how data should be structured.",
            "data\_flow": "DETAILED description of how data flows between operations. What transformations are needed, how to pass context between operations, and how to maintain data consistency.",
            "preprocessing": "DETAILED description of any preprocessing needed before sending data to LLMs (if any).",
            "postprocessing": "DETAILED description of any postprocessing needed after receiving LLM outputs (if any)."
        \}\},
        
        \hspace{2em}"summary": "EXTREMELY DETAILED and comprehensive summary of the framework design, operation structure, LLM migration strategy, and how to solve this task using LLMs. Include specific examples of prompt designs, operation flows, and architectural decisions."
        
    \}\}

    \end{tcolorbox}
    
    \caption{Prompt for Strategy\_C Analysis.}
    \label{fig:prompt for strategy_c analysis}
\end{figure*}

\begin{figure*}[htbp]
    \centering
    \begin{tcolorbox}[
        enhanced,
        colframe=Salmon!90!Black,          
        colback=Salmon!20,         
        coltitle=white,         
        fonttitle=\large\bfseries, 
        title={Prompt for Strategy\_D Analysis}, 
        halign title=left,      
        fontupper=\footnotesize,       
        boxrule=1pt,            
        arc=3mm,                
        boxsep=2pt,
        left=8pt,               
        right=8pt,              
        top=4pt,
        bottom=4pt
    ]
    
    \textbf{System\_prompt:} 
    
    You are an expert evaluator and metrics analyst. Your task is to analyze documents about evaluation metrics, benchmarks, and assessment methods related to a specific task, and provide insights on evaluation approaches and standards.
        
    \textbf{User\_prompt:}
    
    Task Description (from task\_keywords thinking): <task\_thinking>
    
    Strategy: <strategy\_name>
    
    Documents Retrieved: <files\_text>
    
    \textbf{IMPORTANT:} 
    
    Please provide EXTREMELY DETAILED and COMPREHENSIVE analysis. The more detailed, the better. Include specific metric definitions, detailed evaluation procedures, step-by-step assessment workflows, and thorough explanations.
    
    Your task:
    
    1. Analyze all the documents above and identify the evaluation metrics, benchmarks, and assessment methods they discuss. Be very specific and detailed about each metric and method.
    
    2. Summarize the key evaluation information relevant to solving this task. Provide EXTREMELY DETAILED descriptions, including but not limited to:
    
       \hspace{2em}\textbf{Standard evaluation metrics and their definitions:} Provide DETAILED definitions for each metric, including mathematical formulas, calculation methods, interpretation guidelines, and specific use cases. Include examples and detailed explanations.
       
       \hspace{2em}\textbf{Benchmark datasets and evaluation protocols:} Detail each dataset with size, format, structure, quality, and provide DETAILED evaluation protocols with step-by-step procedures, data splits, evaluation criteria, and complete assessment workflows. Be extremely thorough.
       
       \hspace{2em}\textbf{Assessment methodologies and procedures:} Provide DETAILED, step-by-step assessment workflows with all stages, evaluation criteria, scoring methods, and complete procedures. Include specific examples and detailed explanations.
       
       \hspace{2em}\textbf{Performance standards and baselines:} Detail performance benchmarks with specific numbers, comparison methods, baseline implementations, and detailed explanations of what constitutes good performance. Be comprehensive.
       
       \hspace{2em}\textbf{Evaluation best practices and guidelines:} Provide detailed best practices with specific guidelines, common mistakes to avoid, validation procedures, and detailed explanations. Include real-world examples.
       
       \hspace{2em}\textbf{Metrics interpretation and analysis methods:} Detail how to interpret each metric, what values indicate good/bad performance, statistical analysis methods, and detailed interpretation guidelines. Be thorough.
    
    3. Provide a structured summary that clearly explains:
    
       \hspace{2em}What evaluation metrics and benchmarks are covered in these documents (with detailed descriptions)
       
       \hspace{2em}What specific evaluation information is needed to assess task performance (be very specific and detailed)
       
       \hspace{2em}How these evaluation approaches relate to the task requirements (provide detailed connections and relationships)
    
    Remember: 
    
    The more detailed and comprehensive your analysis, the better. Include specific metric definitions, detailed evaluation procedures, step-by-step workflows, and thorough explanations throughout.
    
    Please provide a comprehensive and well-structured analysis in JSON format:
    
    \{\{
    
        \hspace{2em}"evaluation\_metrics": ["detailed metric1 with definition and formula", "detailed metric2 with definition and formula", ...],
        
       \hspace{2em} "evaluation\_information": \{\{
            
            \hspace{4em}"standard\_metrics": ["detailed metric1 with calculation method", "detailed metric2 with calculation method", ...],
            
            \hspace{4em}"benchmark\_datasets": ["detailed dataset1 with protocol", "detailed dataset2 with protocol", ...],
            
            \hspace{4em}"assessment\_methodologies": "DETAILED step-by-step assessment workflow with all stages, criteria, scoring methods, and complete procedures. Be extremely thorough.",
            
            \hspace{4em}"performance\_standards": "DETAILED performance benchmarks with specific numbers, comparison methods, baselines, and explanations. Be comprehensive.",
            
            \hspace{4em}"evaluation\_best\_practices": "DETAILED best practices with guidelines, common mistakes, validation procedures, and explanations. Include examples. Be comprehensive.",
            
            \hspace{4em}"metrics\_interpretation": "DETAILED interpretation guidelines with analysis methods, value meanings, and statistical considerations. Be thorough."
        
        \hspace{2em}\}\},
        
        \hspace{2em}"summary": "EXTREMELY DETAILED and comprehensive summary of the evaluation and metrics knowledge, including all metrics, detailed procedures, and thorough explanations..."
        
    \}\}

    \end{tcolorbox}
    
    \caption{Prompt for Strategy\_D Analysis.}
    \label{fig:prompt for strategy_d analysis}
\end{figure*}

\begin{figure*}[htbp]
    \centering
    \begin{tcolorbox}[
        enhanced,
        colframe=Salmon!90!Black,          
        colback=Salmon!20,         
        coltitle=white,         
        fonttitle=\large\bfseries, 
        title={Prompt for Node Template}, 
        halign title=left,      
        fontupper=\footnotesize,       
        boxrule=1pt,            
        arc=3mm,                
        boxsep=2pt,
        left=8pt,               
        right=8pt,              
        top=4pt,
        bottom=4pt
    ]
    
    
    def \{node\_name\}(self, input\_data): \\
    \hspace{2em}""" \\
    \hspace{2em}node\_id: \{node\_id\} \\
    \hspace{2em}node\_type: \{node\_type\} \\
    \hspace{2em}description: \{description\} \\
    \hspace{2em}dependencies: \{dependencies\} \\
    \hspace{2em}input: \{input\} \\
    \hspace{2em}output: \{output\} \\
    \hspace{2em}""" \\
    \hspace{2em}\# ---------- Step 1: Process the input data\\
    \hspace{2em}\# input\_data is a dictionary with the keys as the input names and the values as the input values \\
    \hspace{2em}\# First, extract the input values from the input\_data dictionary \\
    \hspace{2em}\# Fill your code here \\
    \\
    \\
    \hspace{2em}\# ---------- Step 2: Implement the node logic for one of the node types (LLM\_Generator, Retrieval\_RAG) \\
    \hspace{2em}\# Second, for LLM\_Generator nodes, use LLMs to process the input data \\
    \hspace{2em}\# For example, define the system prompt and user prompt: \\
    \hspace{2em}\# node\_messages = [ \\
    \hspace{4em}\# \{"role": "system", "content": System Prompt from the prompt\_template\}, \\
    \hspace{4em}\# \{"role": "user", "content": User Prompt from the prompt\_template (embed the input values) + Constraints from the constraints field\}, \\
    \hspace{2em}\# ] \\
    \hspace{2em}\# Then, call the LLM to get the output. If there are multiple LLM calls, you should call the LLMs according to the logic\_description field. \\
    \hspace{2em}\# For example, use self.llm\_client.chat(node\_messages, response\_format='json\_object') to get the json format output \\
    \hspace{2em}\# Use self.llm\_client.chat(node\_messages, response\_format='normal') to get the normal text output \\
    \hspace{2em}\# Fill your code here \\
    \\
    \\
    \hspace{2em}\# For Retrieval\_RAG nodes, find the information that this node needs to retrieve from the logic\_description \\
    \hspace{2em}\# Use self.search\_engine.multi\_turn\_search(information needed to retrieve) to get the retrieved context \\
    \hspace{2em}\# Based on the retrieved context, use the summarization prompt template (marked as User Prompt: in the prompt\_template) to summarize the retrieved context \\
    \hspace{2em}\# Use self.llm\_client.chat(node\_messages, response\_format='json\_object') to get the json format output \\
    \hspace{2em}\# Use self.llm\_client.chat(node\_messages, response\_format='normal') to get the normal text output \\
    \hspace{2em}\# Fill your code here \\
    \\
    \\
    \hspace{2em}\# ---------- Step 3: Collect the output\\
    \hspace{2em}\# Finally, collect the output into a dictionary with the keys as the output names and the values as the output values \\
    \hspace{2em}\# Fill your code here \\
    \\
    \\
    \hspace{2em}return output\_data \\

    \end{tcolorbox}
    
    \caption{Prompt for Node Template.}
    \label{fig:prompt for node template}
\end{figure*}

\begin{figure*}[htbp]
    \centering
    \begin{tcolorbox}[
        enhanced,
        colframe=Salmon!90!Black,          
        colback=Salmon!20,         
        coltitle=white,         
        fonttitle=\large\bfseries, 
        title={Prompt for Node Generation Part 1}, 
        halign title=left,      
        fontupper=\footnotesize,       
        boxrule=1pt,            
        arc=3mm,                
        boxsep=2pt,
        left=8pt,               
        right=8pt,              
        top=4pt,
        bottom=4pt
    ]
    
    \textbf{System\_prompt:}
    
        You are an expert system architect and multi-agent system designer. Your task is to design a complete pipeline of nodes (operators) to solve a specific task based on the task description and strategy analysis. You must carefully identify every step the task requires and create a corresponding node for each, do not omit necessary steps. You may ONLY use two types of nodes: \textbf{LLM\_Generator} (call LLM to do reasoning/generation) and \textbf{Retrieval\_RAG} (use search engine for RAG). All verification, validation, parsing, and format-checking must be implemented via the LLM (by writing clear requirements and rules in the prompt\_template so the LLM performs checks and outputs the correct format). Do NOT write code to verify, parse, or validate LLM outputs, use the LLM to do it. Each node must follow the provided node definition structure and work together to form a complete solution pipeline.

    \textbf{User\_prompt:}
    
    Task Description: 
    <task\_thinking>
    
    <task\_samples\_section>
    
    Strategy Analysis:
    <strategy\_analysis>
    
    The code template for all nodes is (Only use for the all\_code field in the node definition):
    <code\_template>
    
    \textbf{IMPORTANT:} 
    Design a pipeline using ONLY two node types. Each node must follow the node definition structure above.
    
    \textbf{STRICT RULES:}
    
    \hspace{2em}Allowed node types: LLM\_Generator and Retrieval\_RAG.
    
    \hspace{2em}Verification and parsing via LLM, NOT code: Any need for verification (e.g. format check, validity check, number validation), parsing (e.g. extracting structured data from text), or fixing malformed output must be implemented by the LLM: put the rules and expected output format in the prompt\_template (System Prompt / User Prompt) so that the LLM performs the checks and returns well-formed output. Do NOT write Python code to validate (e.g. json.loads, try/except, re.match) or parse LLM responses, if output might be messy, add instructions in the prompt or add another LLM\_Generator node that asks the LLM to clean/validate and re-output.
    
    \hspace{2em}For calculations or deterministic steps, use an LLM\_Generator node: ask the LLM to perform the reasoning and output the result in the required format; do not use code.
    
    Your task:
    
    1. Analyze the task and strategy analysis to understand:
       What is the overall task that needs to be solved?
       What background knowledge, architectural patterns, and evaluation metrics are available?
       List exhaustively all operations and workflow steps the task requires (e.g. input parsing, fact extraction, knowledge retrieval, reasoning, validation, synthesis, final answer formatting). Do not skip or merge steps mentally, write them down. Each of these should eventually map to at least one node.
    
    2. Design a complete pipeline of nodes using ONLY LLM\_Generator and Retrieval\_RAG:
    
       \hspace{2em}For each step you identified above, create a corresponding node. Do not generate too few nodes: the pipeline must have enough nodes to cover the entire task from input to final output. If the task typically needs e.g. extraction → retrieval → reasoning → synthesis → formatting, you must have nodes for each (or clearly combined in a justified way).
       Break down the task into logical steps; each step is either (a) call LLM to do something, or (b) use search engine to retrieve then LLM to summarize/use.
       
       \hspace{2em}Before finalizing the node list, double-check: Is there a node that handles retrieval if the task needs external knowledge? Is there a node that produces the final answer in the required format? Are there nodes for every distinct logical phase (e.g. understand input, gather context, reason, output)? Add nodes if any required step is missing.
       
       \hspace{2em}Nodes are connected through dependencies (dependencies field).
       
       \hspace{2em}Do NOT add any node that would require custom Python code (e.g. no "Calculator Tool", "Validator Tool", "Parser Tool" as Python code). Use LLM\_Generator for such roles if needed.
    
    3. For each node, provide complete information following the node definition:
    
       \hspace{2em}node\_name: A descriptive name (e.g., "xx\_Agent").
       
       \hspace{2em}node\_type: One of [LLM\_Generator, Retrieval\_RAG].
       
       \hspace{2em}description: Summary of the node's role in the pipeline.
       
       \hspace{2em}dependencies: List of upstream node names that this node depends on.
       
       \hspace{2em}input: What information this node reads from inputs (be specific based on task samples).
       
       \hspace{2em}output: What this node produces (be specific about output format).
       
       \hspace{2em}constraints: Global constraints this node must comply with (from task requirements).
       
       \hspace{2em}implementation:
         logic\_description: Detailed description of the implementation logic (no code; describe what the node does in terms of LLM calls and/or search + LLM).
         prompt\_template: (For both node types) MUST provide complete, detailed prompt content: System Prompt (marked as "System Prompt:") and User Prompt (marked as "User Prompt:") with placeholders. Be specific and include examples.
         tools\_needed: For Retrieval\_RAG nodes use ["Search"]; for LLM\_Generator use [].
         Do NOT include "code\_snippet". Omit it or set to null.
         
       \hspace{2em}all\_code: Minimal runnable code only: (1) Read inputs from input\_data. (2) For LLM\_Generator: fill the prompt\_template with input values and call self.llm\_client.chat(node\_messages, response\_format=...). (3) For Retrieval\_RAG: build search query from inputs, call self.search\_engine.multi\_turn\_search(query), then fill prompt\_template with retrieved context and call self.llm\_client.chat. (4) Return output\_data dict. Do NOT add code that verifies, parses, or validates the LLM response (no json.loads, re, try/except for parsing, no format checks)—all verification/parsing is done by the LLM via the prompt.

    \end{tcolorbox}
    
    \caption{Prompt for Node Generation Part 1.}
    \label{fig:prompt for node generation part 1}
\end{figure*}

\begin{figure*}[htbp]
    \centering
    \begin{tcolorbox}[
        enhanced,
        colframe=Salmon!90!Black,          
        colback=Salmon!20,         
        coltitle=white,         
        fonttitle=\large\bfseries, 
        title={Prompt for Node Generation Part 2}, 
        halign title=left,      
        fontupper=\footnotesize,       
        boxrule=1pt,            
        arc=3mm,                
        boxsep=2pt,
        left=8pt,               
        right=8pt,              
        top=4pt,
        bottom=4pt
    ]

    \hspace{2em}\textbf{CRITICAL:}
    
       \hspace{2em}Verification and parsing LLM's job: If a node needs to ensure valid JSON, correct format, or validated numbers, write these requirements in the prompt\_template (e.g. "Output only valid JSON.", "Validate each amount and output the approved breakdown."). Do NOT implement verification or parsing in all\_code (no json.loads, re, or try/except to fix LLM output). Use the LLM to do verification and output clean results.
       
       \hspace{2em}LLM\_Generator nodes: Provide full System Prompt and User Prompt in prompt\_template; put any validation/format rules there. all\_code must only: extract inputs, build node\_messages from prompt\_template, call self.llm\_client.chat, return \{\{output\_key: response\}\}. No code that parses or validates the response.
       
       \hspace{2em}Retrieval\_RAG nodes: logic\_description must state what to retrieve and how to summarize. prompt\_template must include System Prompt and User Prompt; use a placeholder like \{\{retrieved\_context\}\} or \{\{retrieved\_chunks\}\} for the search result. all\_code must only: build query from inputs, call self.search\_engine.multi\_turn\_search(query), build node\_messages from prompt\_template with retrieved content, call self.llm\_client.chat, return output. No code that parses or validates the response.
       
       \hspace{2em}Retrieval\_RAG: Design so you do NOT retrieve the question itself; retrieve only related knowledge (e.g. laws, case law, background) needed to answer. State this in logic\_description and prompt\_template.
    
    4. Design principles:
    
       \hspace{2em}Use only LLM\_Generator and Retrieval\_RAG.
       
       \hspace{2em}Completeness over brevity: Ensure the pipeline has enough nodes for the task. List all logical steps the task requires (from task description and strategy analysis), then create one node (or more) for each step. When in doubt, add a dedicated node rather than overloading one node with multiple responsibilities. Too few nodes often lead to incomplete or poor results.
       
       \hspace{2em}Each node has a single responsibility. Dependencies form a DAG.
       
       \hspace{2em}Use LLM\_Generator for reasoning, generation, extraction, validation, and any step that would otherwise need "code" (e.g. ask LLM to output structured JSON or numbers).
       
       \hspace{2em}Use Retrieval\_RAG when external knowledge retrieval (search) is needed, then LLM to summarize or use the retrieved context.
    
    5. Output format:
    
       Provide a JSON object with this structure:
       
       \{\{
       
           \hspace{2em}"pipeline\_description": "Overall description of the pipeline and how nodes work together",
           
           \hspace{2em}"nodes": [
               \{\{
               
                   \hspace{4em}"node\_name": "...",
                   
                   \hspace{4em}"node\_type": "LLM\_Generator or Retrieval\_RAG only",
                   
                   \hspace{4em}"description": "...",
                   
                   \hspace{4em}"dependencies": ["..."],
                   
                   \hspace{4em}"input": ["..."],
                   
                   \hspace{4em}"output": ["..."],
                   
                   \hspace{4em}"constraints": "...",
                   
                   \hspace{4em}"implementation": \{\{
                       "logic\_description": "...",
                       "prompt\_template": "...",
                       "tools\_needed": ["Search"] for Retrieval\_RAG, [] for LLM\_Generator
                   \}\},
                   
                   \hspace{4em}"all\_code": "Minimal code only: input extraction, then LLM call(s) or search+LLM, then return output\_data. No verification/parsing blocks."
                   
               \hspace{2em}\}\}, ...],
           
           \hspace{2em}"Connections": "Complete Python code for def execute\_pipeline(self, initial\_input\_data): ... Execute nodes in dependency order; collect inputs from initial\_input\_data or results; call self.NodeName(input\_data); store outputs; return final output. Import json if needed."
           
       \}\}
    
    Remember:
    
    \hspace{2em}Carefully check that the task needs are fully covered by nodes: Before outputting, verify you have a node for every required step (e.g. input understanding, retrieval if needed, reasoning, synthesis, final answer). The number of nodes should be sufficient to solve the task completely—do not output a pipeline with too few nodes.
    
    \hspace{2em}Use only LLM\_Generator and Retrieval\_RAG.
    
    \hspace{2em}All verification and parsing must be done by the LLM: write rules and output-format requirements in the prompt\_template; do not write code to verify or parse LLM output (no json.loads, re, try/except for validation/parsing in all\_code).
    
    \hspace{2em}all\_code must be minimal: read input -> (LLM call or search+LLM) -> return output. No code that checks or parses the LLM response.
    Dependencies must form a valid DAG. Use task samples to align input/output formats.
    
    \hspace{2em}For "Connections": generate the pipeline execution function that runs nodes in dependency order and passes data correctly.

    \end{tcolorbox}
    
    \caption{Prompt for Node Generation Part 2.}
    \label{fig:prompt for node generation part 2}
\end{figure*}

\begin{figure*}[htbp]
    \centering
    \begin{tcolorbox}[
        enhanced,
        colframe=Salmon!90!Black,          
        colback=Salmon!20,         
        coltitle=white,         
        fonttitle=\large\bfseries, 
        title={Prompt for Node Optimization}, 
        halign title=left,      
        fontupper=\footnotesize,       
        boxrule=1pt,            
        arc=3mm,                
        boxsep=2pt,
        left=8pt,               
        right=8pt,              
        top=4pt,
        bottom=4pt
    ]
    
    \textbf{System\_prompt:} 
    
    You are an expert system optimizer and code reviewer. Your task is to analyze a node in a multi-agent pipeline that has the lowest reward and optimize its internal structure to improve performance. All optimizations must be achieved via the LLM. You may: (1) improve existing LLM prompts, (2) introduce new LLM calls where needed, (3) optimize how multiple LLM calls within the same node communicate and interact—e.g. what is passed between calls, in what format, in what order, and how results are aggregated. Do NOT add Python code for rules, regex, normalization, or filtering—fix shortcomings by prompt engineering or by adding/adjusting LLM calls and their communication, not by code.

    \textbf{User\_prompt:}
    \textbf{Question:} <question>
    
    \textbf{Expected Answer:} <answer>
    
    \textbf{\# Node to Optimize}
    
    Node Name: {node\_name}
    
    Node Type: {node\_type}
    
    Node Description: {node\_description}
    
    Node Reward: {node\_reward} (This is the lowest reward, indicating poor performance)
    
    Node Position: Node {node\_index + 1} in the pipeline
    
    \textbf{\# Current Node Implementation}
    
    Implementation Details:
    \{json.dumps(node\_implementation, ensure\_ascii=False, indent=2)\}
    
    Current Code:
    \{node\_all\_code\}
    
    \textbf{\# Pipeline Context}
    
    All Intermediate Outputs (to understand the data flow; when multiple samples exist, each [Sample N] block's node outputs correspond to the [Sample N] Question/Answer in Task Context above):
    \{intermediate\_context\}
    
    \textbf{\# Analysis Task}
    
    Based on the question, expected answer, and the intermediate outputs from all nodes, analyze why this node has the lowest reward and provide optimization suggestions.
    Analysis Steps:
    
    1. Identify the Problem: 
       What is the node's current output? (from intermediate\_outputs)
       How does it differ from what's expected?
       What specific issues are causing the low reward?
    
    2. Root Cause Analysis:
       Is the prompt (for LLM\_Generator/Retrieval\_RAG) clear and specific enough?
       Are the LLM calls structured optimally? If the node has multiple LLM calls, is the communication between them effective—e.g. is the handoff from one call to the next clear, in a good format, and in the right order? Are there missing or redundant steps?
       Is the implementation handling all cases correctly?
       Is the retrieval (for Retrieval\_RAG) getting relevant information?
       Are there any logical errors or missing validations?
    
    3. Optimization Strategy:
       \{optimization\_focus\}
    
    \textbf{CRITICAL:} Fix shortcomings via LLM, not code: You may (1) improve existing prompts, (2) introduce new LLM calls (e.g. a refinement or validation step), (3) optimize inter-LLM communication when a node has multiple LLM calls—e.g. clarify what each call receives from the previous one, improve the handoff format in the prompt, reorder or add calls so the flow is clearer. Do NOT add Python code for rule-based checks, regex, normalization, or filtering. The code should remain minimal: prepare inputs → call LLM(s), passing outputs between calls as needed → return output.
    
    \textbf{\# Output Format}
    
    Provide a JSON object with the following structure:
    
    \{\{
    
        \hspace{2em}"analysis": \{\{
            "problem\_identification": "Detailed description of what's wrong with the current node",
            "root\_cause": "Analysis of why the node is performing poorly",
            "optimization\_strategy": "Specific strategy to improve the node"
        \}\},
        
        \hspace{2em}"optimized\_implementation": \{\{
            "prompt\_template": "Updated prompt template (marked as System Prompt: and User Prompt:). Keep original if no changes needed.",
            "tools\_needed": "Updated tools\_needed (for Retrieval\_RAG nodes). Keep original if no changes needed.",
            "logic\_description": "Updated logic description explaining the optimization"
        \}\},
        
        \hspace{2em}"optimized\_all\_code": "Complete updated code for the node following the code\_template structure. MUST be complete and runnable. Output the code in the same format as the original code.
        
        \hspace{2em}"optimization\_explanation": "Detailed explanation of what was optimized and why".
        
    \}\}
    
    \textbf{IMPORTANT:}
    
    Fix any identified problems by improving the prompt, adding or reordering LLM calls, or improving how multiple LLM calls in the same node communicate (what is passed, in what format). Do NOT add Python code for validation, regex, normalization, rule-based filtering, or parsing of LLM output. optimized\_all\_code must stay minimal: get inputs → call LLM(s), passing outputs between calls as needed → return output.
    
    For LLM\_Generator: Improve prompt\_template and/or the number and sequence of LLM calls; if there are multiple calls, ensure each call’s prompt clearly receives and uses the outputs of previous calls. Do not add code to parse or validate responses.
    
    For Retrieval\_RAG: Improve the summarization prompt and query construction; do not add code to filter or normalize retrieved content—instruct the LLM to do it in the prompt.
    
    The optimized\_all\_code MUST be complete and runnable but MUST NOT contain extra validation/parsing/regex/filtering code.
    
    If you add, remove, or reorder LLM calls, or change how they communicate (handoff format/order), explain the reasoning in logic\_description.

    \end{tcolorbox}
    
    \caption{Prompt for Node Optimization.}
    \label{fig:prompt for node optimization}
\end{figure*}

\end{document}